\documentclass{article}

\usepackage[preprint]{neurips_2024}

\usepackage{floatrow}  % For customizing floats
\usepackage{graphicx}  % For including figures
\usepackage{booktabs}  % For better-looking tables
\usepackage{blindtext} % For dummy text
\newfloatcommand{capbtabbox}{table}[][\FBwidth]

\usepackage[utf8]{inputenc} % allow utf-8 input
\usepackage[T1]{fontenc}    % use 8-bit T1 fonts
\usepackage{hyperref}       % hyperlinks
\usepackage{url}            % simple URL typesetting
\usepackage{booktabs}       % professional-quality tables
\usepackage{amsfonts}       % blackboard math symbols
\usepackage{nicefrac}       % compact symbols for 1/2, etc.
\usepackage{microtype}      % microtypography
\usepackage{algorithm}
\usepackage{algpseudocode}
\usepackage{amsmath}
\usepackage{graphicx}
\usepackage[table]{xcolor}
\usepackage{xurl}

\title{Conformal Prediction for Multimodal Regression
}

\author{%
  Alexis Bose, Jonathan Ethier,  Paul Guinand \\
  Communications Research Centre Canada
}

\begin{document}

\maketitle

\begin{abstract}
\textbf{This paper introduces multimodal conformal regression. Traditionally confined to scenarios with solely numerical input features, conformal prediction is now extended to multimodal contexts through our methodology, which harnesses internal features from complex neural network architectures processing images and unstructured text. Our findings highlight the potential for internal neural network features, extracted from convergence points where multimodal information is combined, to be used by conformal prediction to construct prediction intervals (PIs). This capability paves new paths for deploying conformal prediction in domains abundant with multimodal data, enabling a broader range of problems to benefit from guaranteed distribution-free uncertainty quantification.}
\end{abstract}

\section{Background and related work}
The comprehension of the internals of neural networks has begun with efforts towards explainability but little has been done to exploit these internal features. The area of conformal prediction (CP) for regression is no different. The consequence is that rich multimodal input features such as images, unstructured text, and categoricals have not participated in the conformal regression prediction uncertainty quantification process.

CP provides a straightforward framework for constructing statistically rigorous uncertainty intervals for model predictions. Importantly, these intervals are valid regardless of the underlying distribution, providing explicit, non-asymptotic guarantees without depending on specific distributional or model assumptions. By using conformal prediction on any pre-trained model, it is possible to create intervals that consistently contain the true value with a specified coverage level. \citet{angelopoulos2021gentle}. 

CP input assumptions are based on data exchangeability. There is no other assumption about the data distribution or the model. This condition is satisfied for the problems outlined in this paper.

The more developed field of CP classification has done image-to-image uncertainty \citet{pmlr-v162-angelopoulos22a}, and even proposed changing CP regression tasks into classification problems \cite{guha2023conformal}. Despite a comprehensive literature review, including the \textit{Awesome Conformal Prediction} repository \citep{Manokhin_Awesome_Conformal_Prediction_2024}, no findings similar to those proposed by our novel use of internal features in CP for multimodal regression have been uncovered.

The inductive, or split, approach to creating conformal regressors, outlined by \citet{papadopoulos2002inductive}, involves splitting the training data into a proper training set and a calibration set, both of which should accurately represent the overall data distribution. The training set is used to build the regression model, while the calibration set is used to calculate the model's absolute residuals. Conformal predictive systems (CPS) enhance conformal regressors by generating cumulative probability distributions, over potential target values \citet{vovk2020computationally}. This enables the creation of PIs with a specified confidence level, effectively capturing prediction uncertainty while providing a statistically grounded basis for decision-making.

Typically, the more complicated multimodal inputs of images and unstructured text are processed by larger more complicated networks (e.g., convolutional neural networks (CNN), transformers, hybrid networks, etc.). These networks can be compute-intensive. Therefore, building multiple models and having PIs generated in an ensemble fashion may not be an option. 
The inductive implementation of the aforementioned CPSs was the chosen approach as the model is only trained once after a hold-out calibration set is removed from the training set.

Multimodal architectures generally contain an internal combining stage where all the processed inputs come together. These internal inputs will be referred to as internal features within the respective architectures. Conversely, external features are features that are fed to the model at the input layer. The possible advantage with working with internal features is that they have been filtered for significance and weighted for importance which is not true of the input features. This results in a metric suitable for distance-based conformal prediction. 

The major contribution of this work is to address the challenges associated with conformal prediction in the context of multimodal regression tasks involving diverse input data, such as tabular, unstructured text, and image modalities. Traditional conformal prediction methods face significant obstacles when applied directly to such heterogeneous input features in regression problems. However, this paper demonstrates a novel approach that leverages an internal feature layer within the model architecture for multimodal regression. Specifically, we show that this intermediate representation can be effectively utilized to generate a calibration set suitable for conformal prediction in regression scenarios. This methodology extends the applicability of conformal prediction techniques to complex, multimodal regression tasks, thereby enhancing uncertainty quantification capabilities for a wider range of machine learning regression models and diverse data types.

\section{Experiment framework}

A number of multimodal conformal regression problems were investigated using the Conformal Classifiers Regressors and Predictive Systems (CREPES) framework.\footnote{The official code implementation for this paper is available at \url{https://github.com/ic-crc/uncertainty-estimation}.}

The CP training set was utilized to fit the difficulty estimator, which was then applied to the CP calibration set to produce calibrated difficulty estimates. Subsequently, a conformal regressor (CR) model is constructed using the calibration residuals along with these calibrated difficulty estimates. This conformal regressor converts point predictions into prediction intervals at a specified confidence level. To generate prediction intervals for the test set, difficulty estimates are created based on the predicted targets. These test set difficulty estimates, and the confidence level, are then employed by the CR to generate the prediction intervals.

The normalized conformal regressors employed in CREPES implementation of CPSs were: \textit{norm\_std}, which normalizes the standard deviation of the targets from the K-nearest neighbours, and \textit{norm\_targ\_strng}, the Target Strangeness described in \citet{bose2024targstrg}. A different approach to generating PIs of varying sizes involves partitioning the object space into distinct, non-overlapping Mondrian categories \citep{crepes}, where predictions are grouped into bins for categorization. The Mondrian variants employed in the CPSs were: \textit{norm\_std}, \textit{norm\_targ\_strng}, and \textit{norm\_res}, which normalizes on the residual of the K-nearest neighbours.

\section{Experiment preparation}

The first problem centred on predicting Reference Signal Received Power (RSRP). RSRP is an industry-standard metric for measuring the strength of wireless signals received by user devices, such as phones, and it depends on transmitted power and path loss. A model was developed using satellite imagery to correct path loss estimates to predict RSRP. Both multimodal external and internal features were utilized to generate and compare PI performance.

In the second problem, we used a multimodal toolkit \citet{gu-budhkar-2021-package} to confirm the viability of PIs using larger internal features from a Bidirectional Encoder Representations from Transformers (BERT), and Multi-Layer Perceptron (MLP) networks. 

The data preparation for both experiments aimed to have a calibration set size of over 1000, as a set of this size is generally sufficient for most purposes \citet{angelopoulos2021gentle}. Also, the calibration sets were verified to be representative of the training set.

\subsection{DTU Reference Signal Received Power (RSRP) Regression Model}
\subsubsection{Data preparation}
\label{DTU_data_prep}
\begin{figure}[htbp]
\begin{floatrow}
\ffigbox{%
  \includegraphics[width=1\linewidth]{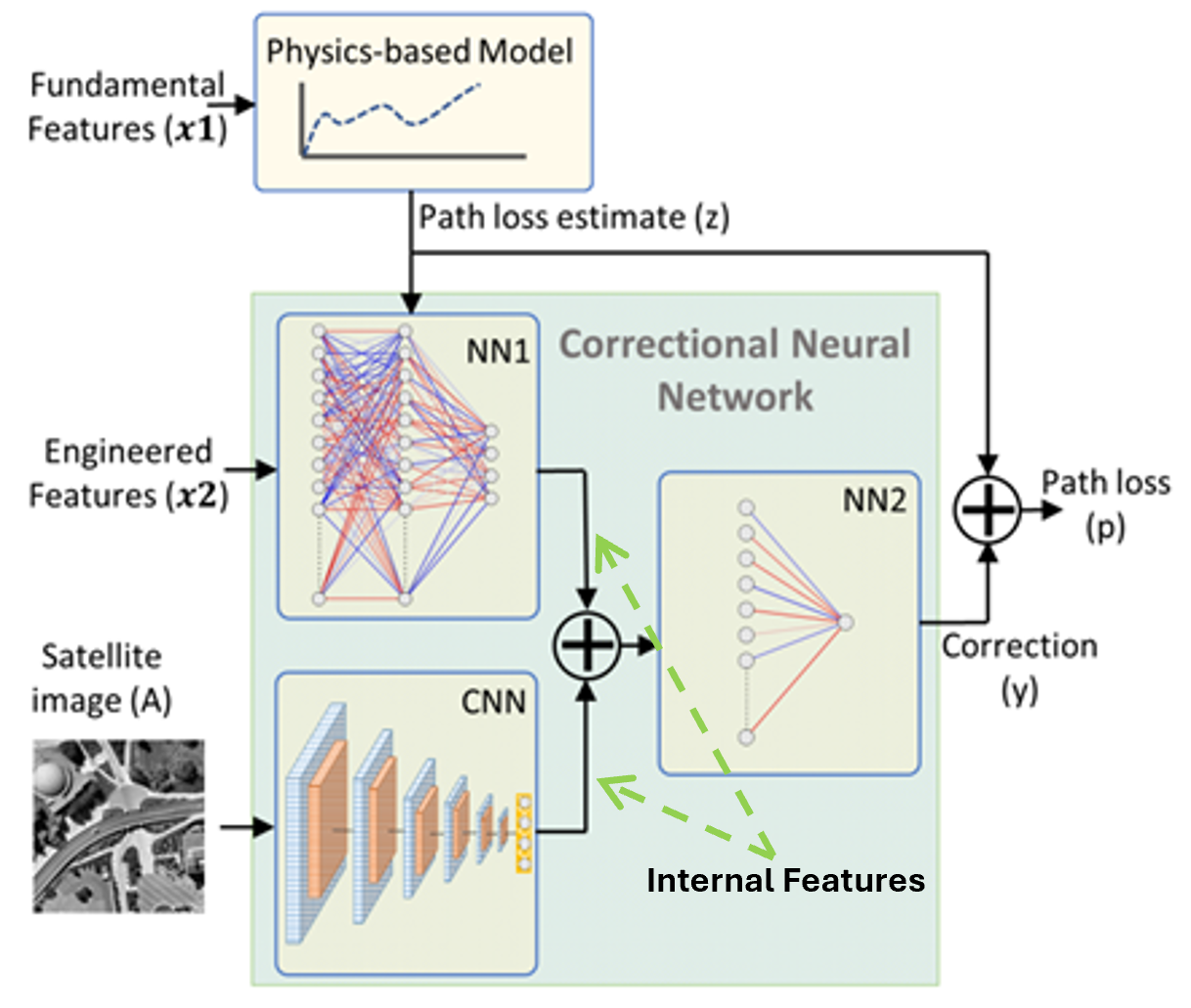}%
}{%
  \caption{DTU RSRP Correctional Neural Network \citet{Thrane2020}, \citet{nguyen2023deep}}
  \label{fig:DTU-RSRP Network Architecture}%
}
\capbtabbox{%
  \begin{tabular}{@{}lll@{}} % Adjusted column widths
    \toprule
    \textbf{Dataset} & \textbf{Count} & \textbf{Description} \\
    \midrule
    train & 35708 & \\
    cal & 3968 & \\
    test & 7114 & \\
    external features & 9 & Lon, Lat, \\
    & & Dist-x,y\\
    & & PCI*\\
    & & path loss \\
    internal image & 4 & CNN output \\
    internal numerical & 4 & NN output \\
    \bottomrule
    \multicolumn{3}{c}{*Physical Cell Identities 64, 65 and 302}
  \end{tabular}
}{%
  \caption{DTU RSRP dataset counts}
  \label{tab:DTU-RSRP_dataset_counts}%
}
\end{floatrow}
\end{figure}

Following the instructions from \citet{Thrane2020}, the Denmark Technical University (DTU) campus RSRP drive test with satellite images was downloaded. The geographically split train and test sets are outlined in \citet{Thrane2020}. The CP calibration set was randomly sampled and removed from the training set, before model training. The models were modified to output their internal features as described in Table \ref{tab:DTU-RSRP_dataset_counts}.
Note that this dataset contains two different frequencies, and as such, creates two distinct clusters in the prediction interval plots. The CP train, calibration and test sets were generated by using the trained model to predict targets and create files with features indicated in Table \ref{tab:DTU-RSRP_dataset_counts}.

\subsubsection{Model architecture}
There are two trained models derived from the DTU RSRP architecture. One is the full model where both the external features and image are used as inputs to model the difference between a path loss estimate and a measured RSRP, referred to as the Correctional Neural Network in Figure \ref{fig:DTU-RSRP Network Architecture}. The second is an image-only model that corrects a path loss into an RSRP. Specifically, the NN1 is removed from Figure \ref{fig:DTU-RSRP Network Architecture}.
The training of each model was done with hyperparameters defined in the code and paper \citet{Thrane2020}. The training took approximately 45 minutes for 20 epochs. The full model had a blind test root mean squared error (RMSE) of 6.95 dB while the image-only model had an RMSE of 14.47 dB. It is important to note the relative difference in RMSE of the underlying models when comparing PIs results.

\subsection{Multimodal Toolkit: Price Regression Model}
\subsubsection{Data preparation}
\label{Multimodal_data_prep}
\begin{figure}[htbp]
\begin{floatrow}
\ffigbox{%
  \includegraphics[width=1\linewidth]{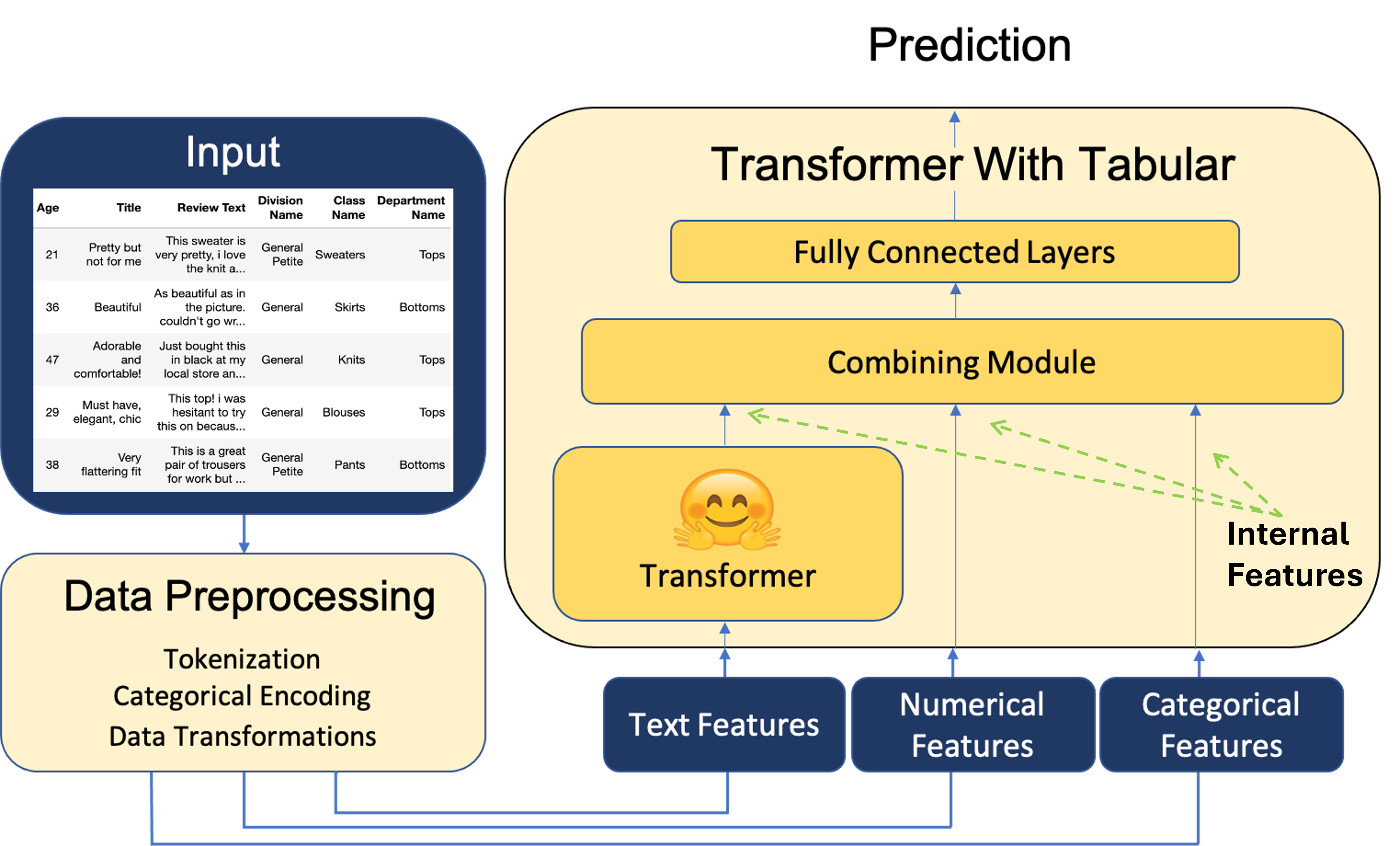}%
}{%
  \caption{Multimodal Toolkit architecture \citet{gu-budhkar-2021-package}}
  \label{fig:Multimodal-toolkit Architecture}%
}
\capbtabbox{%
\begin{tabular}{@{}lll@{}} % Adjusted column widths
    \toprule
    \textbf{Dataset} & \textbf{Count} & \textbf{Description} \\
    \midrule
    train & 9721 & random\\
    calibrate & 1000 &  Airbnb \\
              &       & Melbourne,  \\
    test & 1288 & AUS\\
    \midrule
    external features & 15 & numerical \\
                      &    & [74 categorical*, \\
                      &    & 3 text*] \\
    internal text & 768 & BERT output \\
    internal numerical & 15 & MLP output \\
    internal categorical & 2000 & MLP output \\
    \bottomrule
    \multicolumn{3}{c}{*CP cannot use these features}
\end{tabular}

}{%

  \caption{Multimodal Toolkit dataset counts and descriptions}
  \label{tab:Multimodal-toolkit_dataset_counts}%
}
\end{floatrow}
\end{figure}

The price prediction Melbourne Airbnb Open dataset was contained in the \citet{gu-budhkar-2021-package} Multimodal Toolkit repository.\footnote{The feature names can be found in the dataset's column\_info.json.} The CP calibration set was randomly sampled and removed from the training set, before model training. The model was modified to output the internal features as summarized in Table \ref{tab:Multimodal-toolkit_dataset_counts}. Note that when using CP with external features, rows with NaNs were removed from the dataset. As a result, the test set was reduced to 242 samples, while the calibration set remained at 1000 samples. This is not the case with internal features as there are no NaNs present.

\subsubsection{Model architecture}
The Multimodal Toolkit architecture in Figure \ref{fig:Multimodal-toolkit Architecture} was modified to output the internal features that entered their combining stage. The model was trained using the default combining feature method called \textit{gating\_on\_cat\_and\_num\_feats\_then\_sum}; refer to code and paper \cite{gu-budhkar-2021-package} for more details. The training took approximately one hour.

\section{Experiment compute resources}
The only compute resource used in the following experiments was an AWS g5.4xlarge (16 vCPUs, 64 GiB Memory, a10g GPU), as the intensive compute requirements of both the CNN and the transformer required a GPU.

\section{Experimental results}

The following results were obtained by performing a grid search for two hyperparameters: the number of K-nearest neighbours and the number of Mondrian bins (ranging from 10 to 100 in increments of 10), with the exception of the multimodal experiments that used a maximum Mondrian bin of 60  because there was insufficient data to support a larger bin size. This search was conducted for the five outlined difficulty estimators and five different seeds. The hyperparameters yielding the narrowest PI width were chosen for each estimator. The effective coverage is the percentage of predictions that fall within the PIs. To ensure comparability, results were analyzed at approximately a 90\% effective coverage. 
The experiments requested PIs with a 90\% and 95\% \textit{confidence} and took the means around the effective coverage of interest, In some cases, 90\% effective coverage was not achievable, so no results were recorded. More details can be found in the technical appendix, which contains plots with standard deviation error bars. 

The results tables for each model contain the data features and CP method description, mean PI width and finally the optimal K-nearest neighbours (kNN), and Mondrian bins where applicable.

\subsection{DTU Reference Signal Received Power (RSRP) regression model}
Correcting path loss to predict RSRP using tabular features and satellite images constitutes a multimodal regression problem. Additionally, since conformal prediction intervals have been established for path loss \citet{bose2024targstrg}, RSRP PIs should also be possible.
CP using the full model predictions and internal or external features to generate the PIs had similar results as shown in Table \ref{tab:dtu-rsrp_results}. The use of tabular features seemed to be more informative. Recall that the underlying full model had better RMSE than the image-only model, this aleatoric uncertainty tied to output predictions is also reflected in the mean PI results. However, these results do demonstrate that CP can use internal features, and specifically features of images can be taken into account in the PI quantification. The Target Strangeness method did occasionally outperform in some cases. Refer to Figure \ref{fig:PIs_for_Image-Only_Model_using_Internal_Image_Features} for an image-only example of PIs.

\begin{table}[htbp]
  \caption{DTU RSRP prediction interval results}
  \label{tab:dtu-rsrp_results}
  \centering
  \begin{tabular}{llccc}
    \toprule
    \textbf{Configuration} & \textbf{Mean PI} & \textbf{Effective} & & \\
    \textbf{(data and CP method)} & \textbf{Width [dB]} & \textbf{Coverage [\%]} & \textbf{kNN} & \textbf{Bins}\\
    \midrule
    \multicolumn{5}{l}{\textbf{Tabular-Image Model using External Tabular Features}} \\
    norm\_std & & & & \\
    norm\_targ\_strng & 20.9 & 90.0 & 50 & \\
    \midrule
    \multicolumn{5}{l}{\textit{Mondrians}} \\
    norm\_std & 26.1 & 90.0 & 20 & 60 \\
    norm\_res & & & & \\
    \cellcolor{green!30}norm\_targ\_strng & \cellcolor{green!30}20.4 & \cellcolor{green!30}90.0 & \cellcolor{green!30}40 & \cellcolor{green!30}20 \\
    \midrule
    \multicolumn{5}{l}{\textbf{Tabular-Image Model using Internal Tabular, Image Features}} \\
    norm\_std & 31.7 & 93.5 & 90 & \\
    norm\_targ\_strng & 22.9 & 90.0 & 20 & \\
    \midrule
    \multicolumn{5}{l}{\textit{Mondrians}} \\
    norm\_std & 32.0 & 93.0 & 90 & 50 \\
    norm\_res & & & & \\
    \cellcolor{green!30}norm\_targ\_strng & \cellcolor{green!30}22.4 & \cellcolor{green!30}90.0 & \cellcolor{green!30}20 & \cellcolor{green!30}20 \\
    \midrule
    \multicolumn{5}{l}{\textbf{Tabular-Image Model using Internal Tabular Features}} \\
    norm\_std & 22.3 & 92.0 & 30 & \\
    norm\_targ\_strng & 23.8 & 90.0 & 30 & \\
    \midrule
    \multicolumn{5}{l}{\textit{Mondrians}} \\
    \cellcolor{green!30}norm\_std & \cellcolor{green!30}21.1 & \cellcolor{green!30}90.0 & \cellcolor{green!30}30 & \cellcolor{green!30}10 \\
    norm\_res & & & & \\
    norm\_targ\_strng & 25.6 & 90.0 & 20 & 10 \\
    \midrule
    \multicolumn{5}{l}{\textbf{Tabular-Image Model using Internal Image Features}} \\
    norm\_std & & & & \\
    norm\_targ\_strng & & & & \\
    \midrule
    \multicolumn{5}{l}{\textit{Mondrians}} \\
    \cellcolor{green!30}norm\_std & \cellcolor{green!30}23.5 & \cellcolor{green!30}90.0 & \cellcolor{green!30}20 & \cellcolor{green!30}10 \\
    norm\_res & & & & \\
    norm\_targ\_strng & 23.9 & 90.0 & 10 & 90 \\
    \midrule
    \multicolumn{5}{l}{\textbf{Image-Only Model using Internal Image Features}} \\
    norm\_std & & & & \\
    norm\_targ\_strng & 44.2 & 90.0 & 80 & \\
    \midrule
    \multicolumn{5}{l}{\textit{\textit{Mondrians}}} \\
    \cellcolor{green!30}norm\_std & \cellcolor{green!30}43.6 & \cellcolor{green!30}90.0 & \cellcolor{green!30}70 & \cellcolor{green!30}10 \\
    norm\_res & & & & \\
    norm\_targ\_strng & 44.7 & 91.0 & 30 & 10 \\
    \bottomrule
  \end{tabular}
\end{table}

\subsection{Multimodal Toolkit: Price Regression Model}
\begin{figure}[htbp]
\begin{floatrow}
\ffigbox[\FBwidth]{%
  \includegraphics[width=\linewidth]{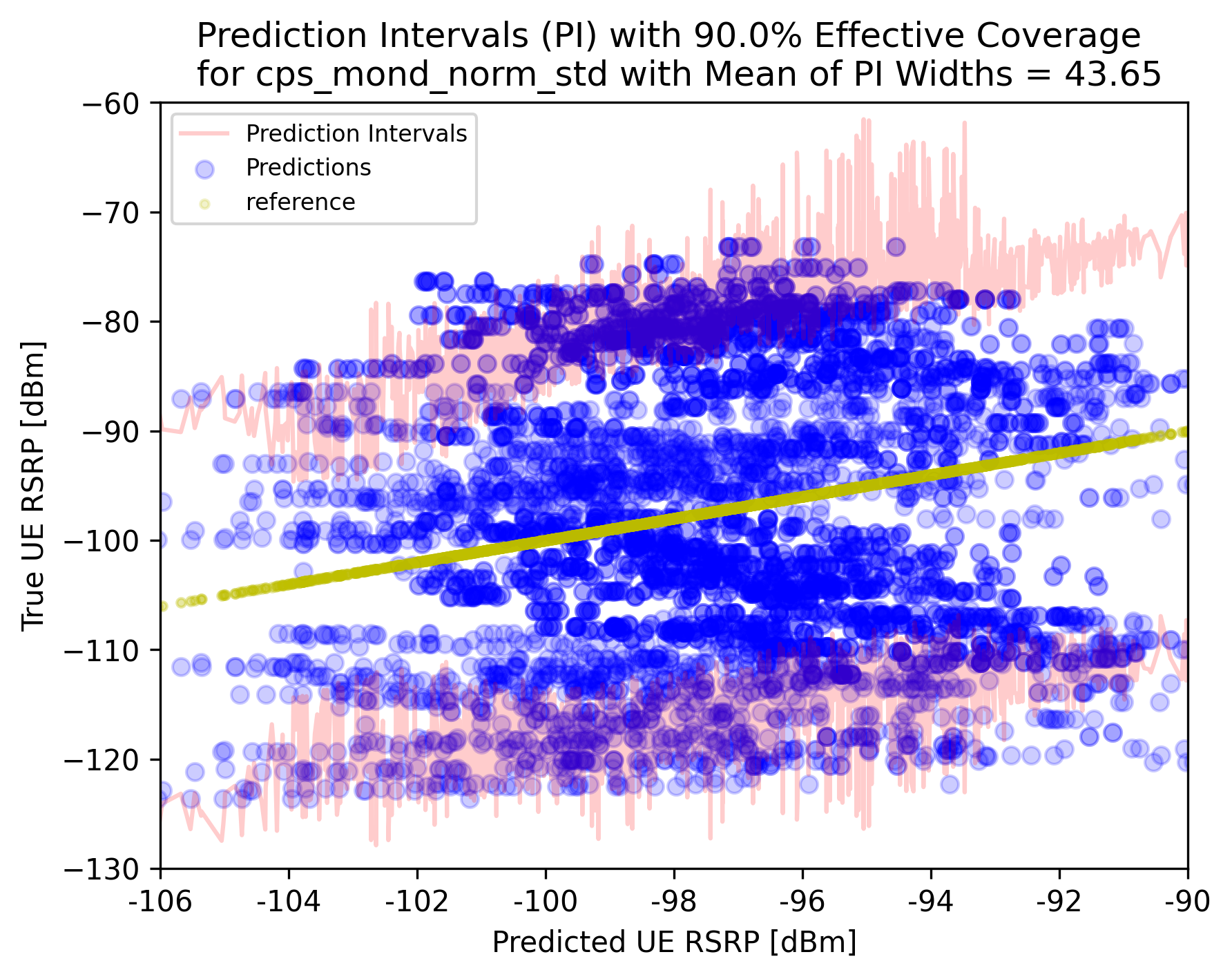}%
}{%
  \caption{PIs from image-only model using internal image features}
  \label{fig:PIs_for_Image-Only_Model_using_Internal_Image_Features}%
}
\ffigbox[\FBwidth]{%
  \includegraphics[width=1\linewidth]{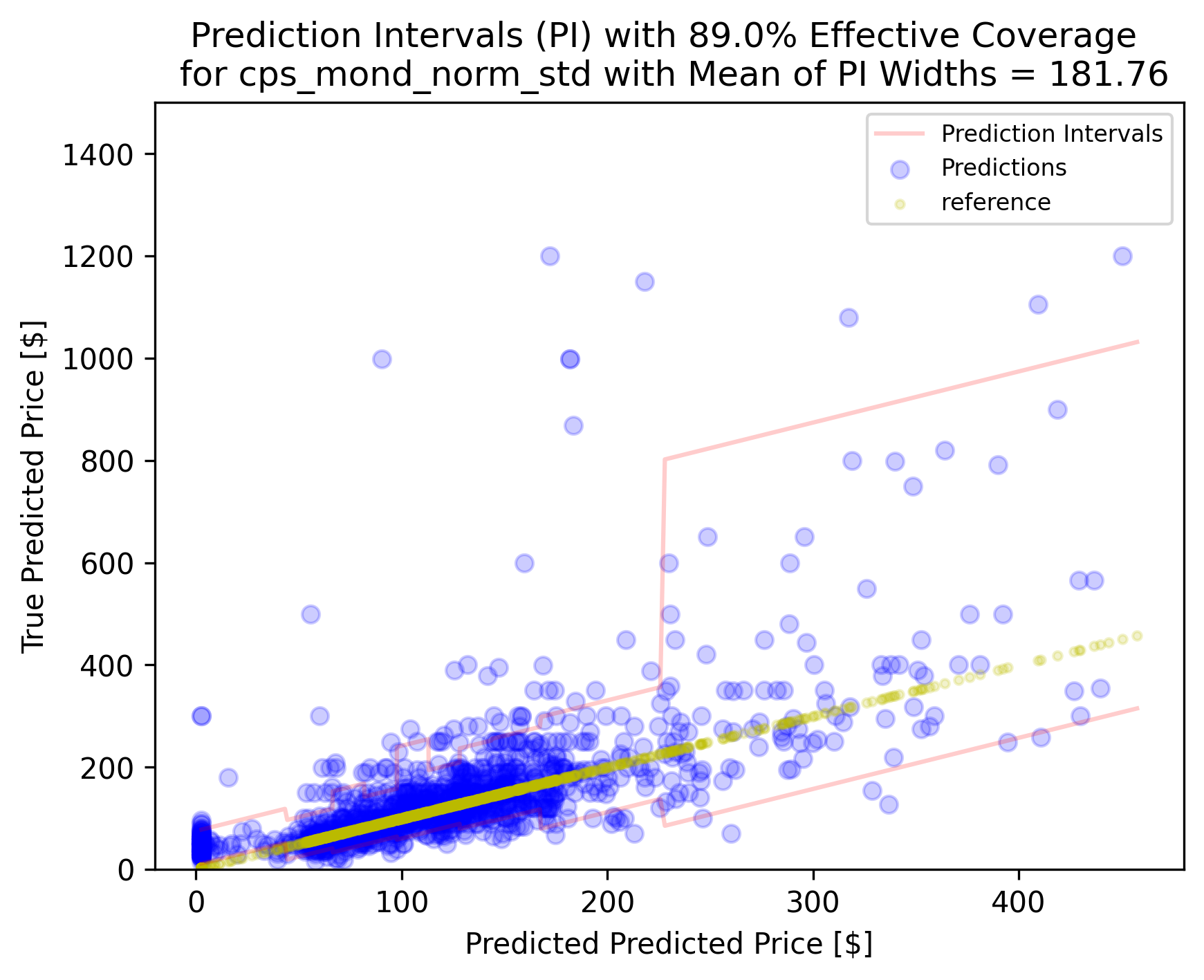}%
}{%
  \caption{PIs from BERT-only internal features}
  \label{fig:PIs_for_Only_BERT_features}%
}
\end{floatrow}
\end{figure}

The results from Table \ref{tab:multimodal-airbnb_results} demonstrate that CP can use internal features from categorical, numerical, and BERT networks to create PIs. This allows CP to take into account categorical and unstructured text. However, the internal features, which all had similar results, did not surpass the mean PI widths of the external features. However, despite their similar performance, the internal features did not yield narrower mean PI widths compared to the external features. The performance of the external features may be attributed to its reduced test set, comprising only 20\% of the samples of other sets due to NaN value elimination. Among the internal features experiments, the BERT-Only was the best. The Target Strangeness difficulty estimator outperformed in all cases except one. An example set of BERT-Only PIs can be seen in Figure \ref{fig:PIs_for_Only_BERT_features}.

\begin{table}[htbp]
  \caption{Multimodal Toolkit rental prediction interval results}
  \label{tab:multimodal-airbnb_results}
  \centering
  \begin{tabular}{llccc}
   \toprule
    \textbf{Configuration} & \textbf{Mean PI} & \textbf{Effective} & & \\
    \textbf{(data and CP method)} & \textbf{Width [\$]} & \textbf{Coverage[\%]} & \textbf{kNN} & \textbf{Bins}\\
    \midrule
    \multicolumn{5}{l}{\textbf{External Numerical Features}} \\
    norm\_std & 193.7 & 91.2 & 20 & \\
    \cellcolor{green!30}norm\_targ\_strng & \cellcolor{green!30}175.1 & \cellcolor{green!30}92.4 & \cellcolor{green!30}60 & \\
    \midrule
    \multicolumn{5}{l}{\textit{Mondrians}} \\
    norm\_std & 217.7 & 90.7 & 10 & 10 \\
    norm\_res & 207.4 & 92.0 & 10 & 10 \\
    norm\_targ\_strng & 182.6 & 91.2 & 50 & 10 \\
    \midrule
    \multicolumn{5}{l}{\textbf{Internal BERT, Categorical, Numerical, and Features}} \\
    norm\_std & 220.8 & 90.2 & 20 & \\
    \cellcolor{green!30}norm\_targ\_strng & \cellcolor{green!30}198.5 & \cellcolor{green!30}91.4 & \cellcolor{green!30}70 & \\
    \midrule
    \multicolumn{5}{l}{\textit{Mondrians}} \\
    norm\_std & 242.6 & 90.0 & 20 & 10 \\
    norm\_res & 232.1 & 90.2 & 10 & 10 \\
    norm\_targ\_strng & 216.3 & 90.8 & 50 & 10 \\
    \midrule
    \multicolumn{5}{l}{\textbf{BERT-Only Features}} \\
    \cellcolor{green!30}norm\_std & \cellcolor{green!30}190.9 & \cellcolor{green!30}90.6 & \cellcolor{green!30}10 & \\
    norm\_targ\_strng & 198.0 & 90.0 & 10 & \\
    \midrule
    \multicolumn{5}{l}{\textit{Mondrians}} \\
    norm\_std & 208.6 & 89.8 & 10 & 10 \\
    norm\_res & 208.6 & 89.8 & 10 & 10 \\
    norm\_targ\_strng & 225.3 & 89.8 & 20 & 10 \\
    \midrule
    \multicolumn{5}{l}{\textbf{Categorical-Only Features}} \\
    norm\_std & 228.6 & 90.0 & 10 & \\
    \cellcolor{green!30}norm\_targ\_strng & \cellcolor{green!30}195.8 & \cellcolor{green!30}91.4 & \cellcolor{green!30}30 & \\
    \midrule
    \multicolumn{5}{l}{\textit{Mondrians}} \\
    norm\_std & 248.3 & 89.8 & 10 & 10 \\
    norm\_res & 245.2 & 89.6 & 10 & 10 \\
    norm\_targ\_strng & 214.6 & 90.4 & 40 & 10 \\
    \midrule
    \multicolumn{5}{l}{\textbf{Numerical-Only Features}} \\
    norm\_std & 198.8 & 89.8 & 10 & \\
    \cellcolor{green!30}norm\_targ\_strng & \cellcolor{green!30}194.5 & \cellcolor{green!30}91.2 & \cellcolor{green!30}20 & \\
    \midrule
    \multicolumn{5}{l}{\textit{Mondrians}} \\
    norm\_std & 213.4 & 89.6 & 10 & 10 \\
    norm\_res & 214.4 & 89.6 & 10 & 10 \\
    norm\_targ\_strng & 212.5 & 90.4 & 10 & 10 \\
    \bottomrule
  \end{tabular}
\end{table}

\newpage

\subsection{Intuition and Rationale}
The rationale behind this paper stems from an important insight into feature representation, emphasizing the role of internal neural network features.

Complex neural networks, particularly models like BERT and CNNs, excel at transforming and equalizing the representations of complex input modalities into a unified and comparable feature space. This internal neural network block effectively normalizes the contributions of disparate inputs, such as unstructured text and images, which originally existed in vastly different scales and dimensionalities.

This approach demonstrates that utilizing internal neural network features makes uncertainty quantification possible in unstructured text-only and image-only regression tasks through conformal prediction. However, in both experiments, while some combinations of internal features achieve equalizing representations, they do not exceed the performance achieved by using external numerical features. This can be explained by the image-only model's greater aleatoric uncertainty, and the reduced CP test set used for the Multimodal Toolkit with external features.

\section{Conclusions}
This paper explored the feasibility of using conformal prediction with different combinations of internal multimodal features with different difficulty estimators. Our findings demonstrate that conformal prediction intervals can be effectively applied to image-only regression problems, and this approach could be extended to other tasks such as image-to-canopy height estimation and crop yield prediction. Moreover, it has been shown that conformal prediction intervals can be effectively constructed using unstructured text alone.
By incorporating CNNs and transformer networks, these results further demonstrate that conformal prediction can scale with more complex networks and the number of internal features. 

\section{Acknowledgements}
I extend my deepest gratitude to my father, Amitava \citet{BoseAmitava}, for his support, and to my grandfather, Asok \citet{BoseAsok} \citet{BoseAsokMemorial2006}, whose scientific legacy inspires me. I also honor my great-grandfather, Akshay Bose, who graduated from the now Visva-Bharati under Rabindranath Tagore in 1908, shaping our family's academic path.

\newpage
\bibliographystyle{plainnat}
\bibliography{Styles/references}

\newpage
\appendix

\section{Appendix}

\subsection{Experimental results}

A grid search was conducted to identify optimal hyperparameters (K-nearest neighbours and Mondrian bins) for five difficulty estimators and five seeds. Although 90\% and 95\% confidence levels were requested, the effective coverage varied based on the data. Hyperparameters leading to the narrowest prediction interval width were selected for each estimator. Mean prediction interval widths were calculated and plotted with error bars representing the standard deviation of errors, assumed to follow a normal distribution.

To ensure comparability, results were analyzed at approximately 90\% effective coverage, requiring means over all configuration results and a separate mean over results with effective coverage exceeding 89\%. For some data-difficulty combinations, comparable mean 90\% effective confidence results were unattainable, so we used means over results with effective coverage greater than 89\%. In some cases, 90\% effective coverage was not achievable, and thus, no results were recorded.

The following experimental configuration and effective coverage plots contain the number of aggregated samples used to calculate the mean and standard deviation.

\subsubsection{DTU RSRP Regression PIs Results}
%Tabular-Image Model using External Tabular Features
\begin{figure}[htbp]
    \centering
    \includegraphics[width=0.7\textwidth]{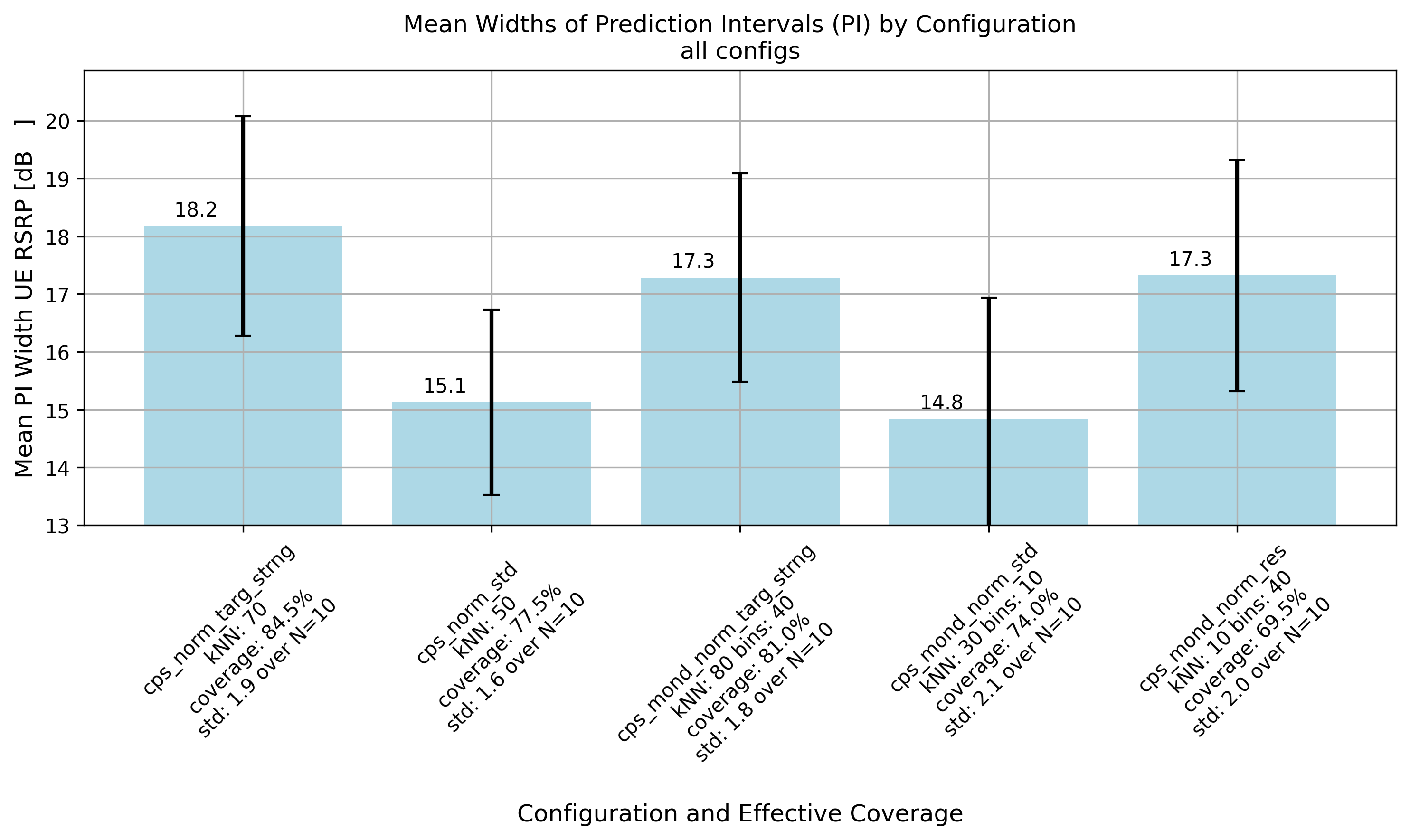}
    \caption{DTU RSRP all experiments: tabular-image model using external tabular features}
\end{figure}

\begin{figure}[htbp]
    \centering
    \includegraphics[width=0.7\textwidth]{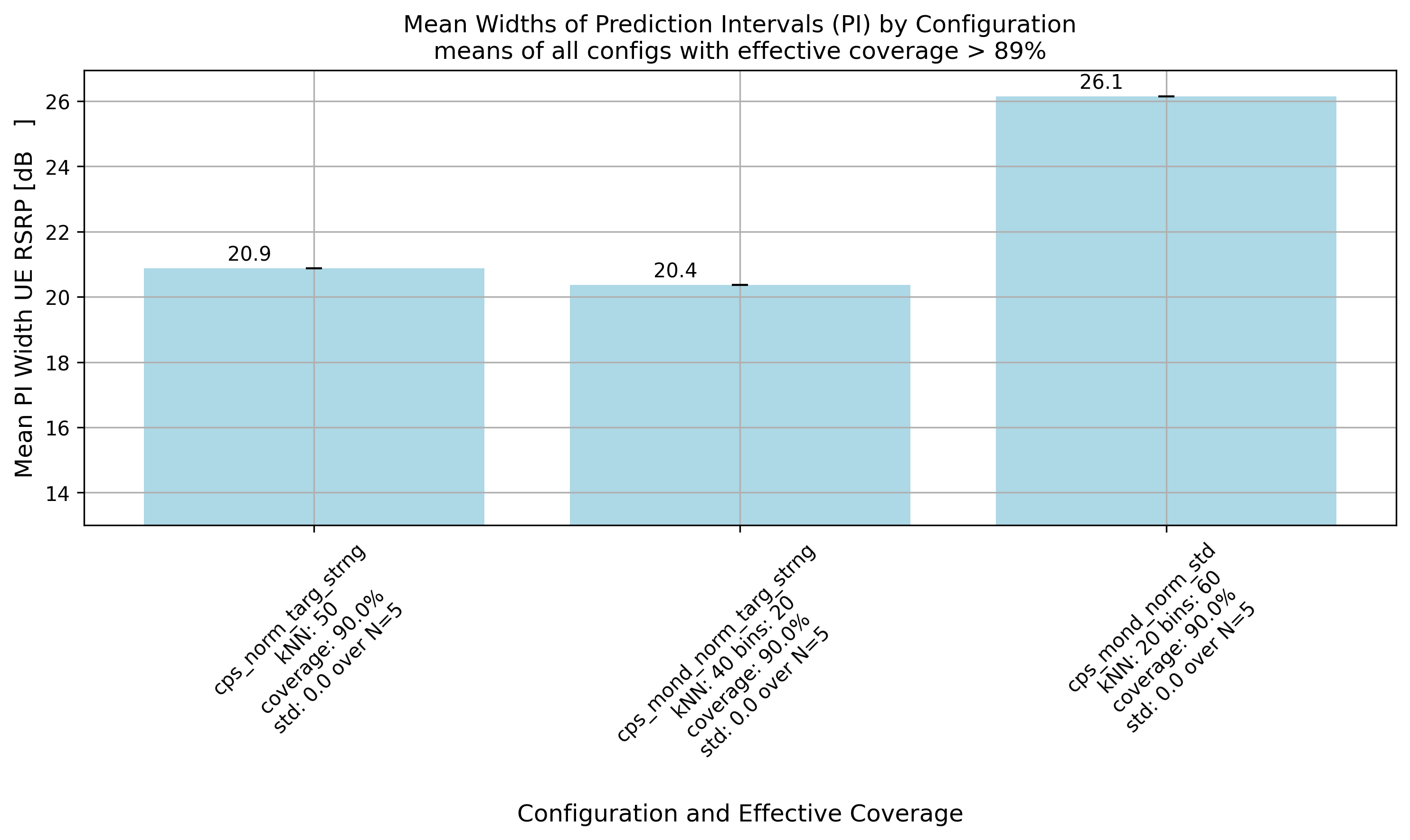}
    \caption{DTU RSRP experiments > 89\% coverage: tabular-image model using external tabular features }
\end{figure}

\begin{figure}[htbp]
    \centering
    \includegraphics[width=0.7\textwidth]{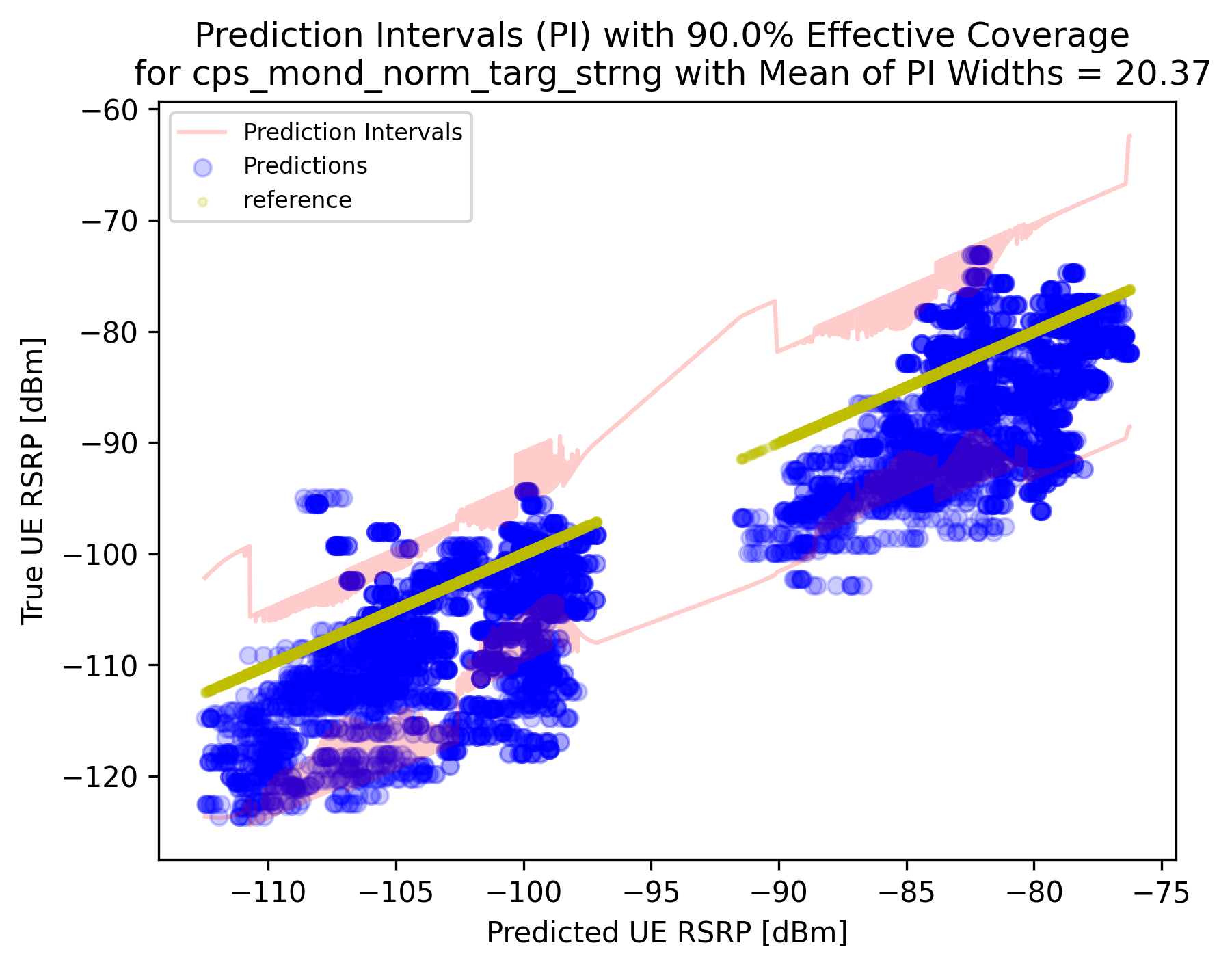}
    \caption{DTU RSRP PIs example: tabular-image model using external tabular features}
\end{figure}

%Tabular-Image Model using Internal Tabular, Image Features

\begin{figure}[htbp]
    \centering
    \includegraphics[width=0.7\textwidth]{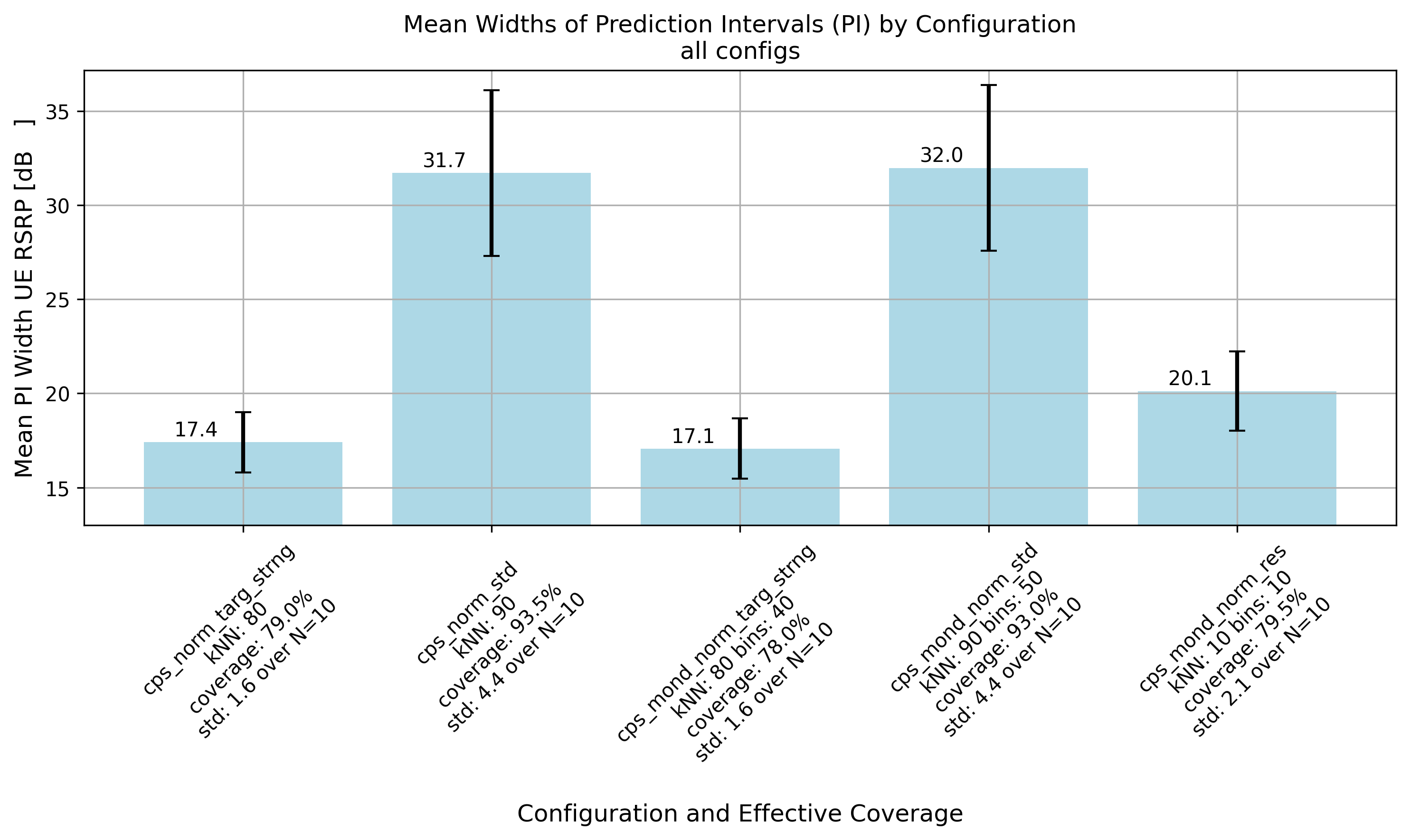}
    \caption{DTU RSRP all experiments: tabular-image model using internal tabular, image features}
\end{figure}

\begin{figure}[htbp]
    \centering
    \includegraphics[width=0.7\textwidth]{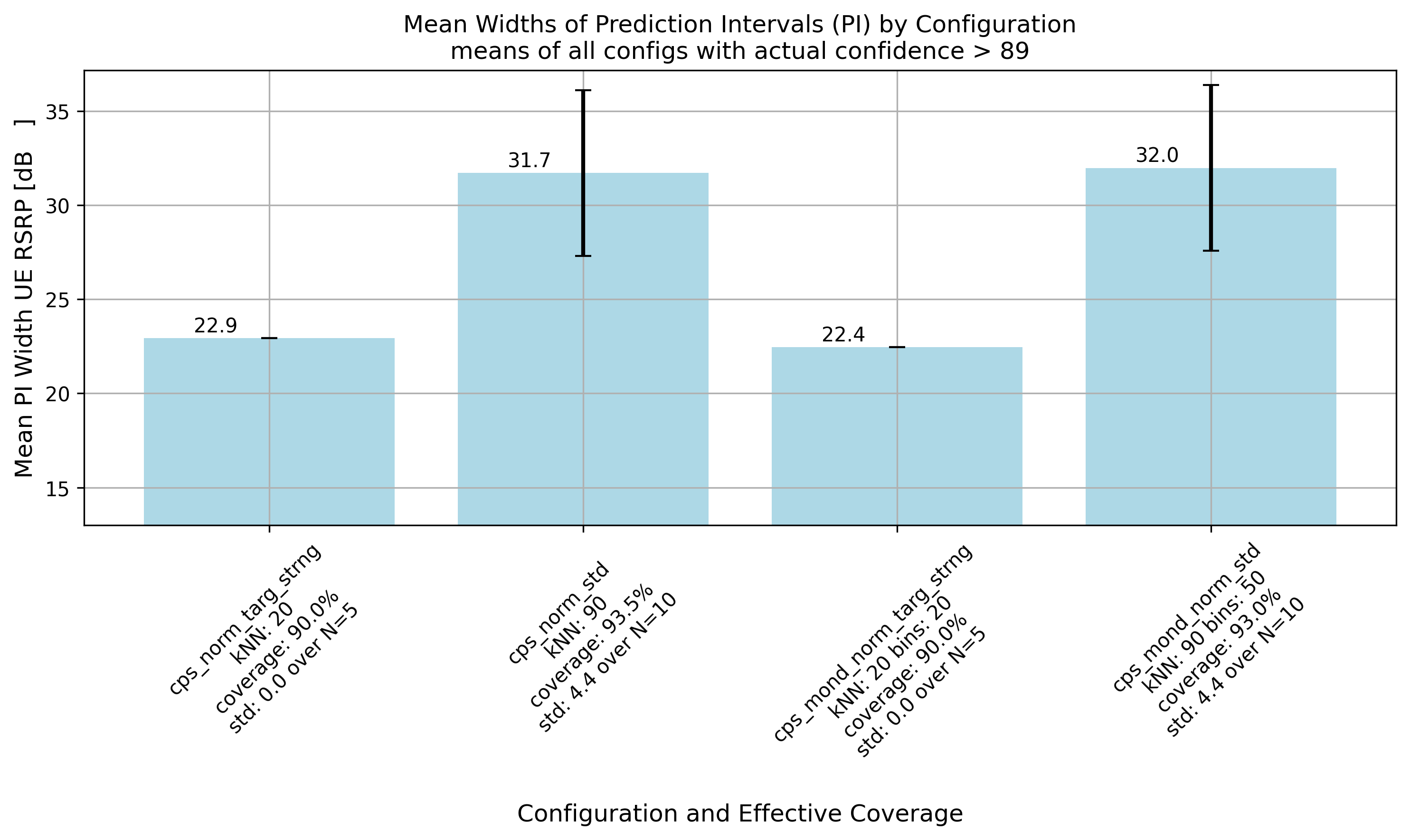}
    \caption{DTU RSRP experiments > 89\% coverage: tabular-image model using internal tabular, image features} 
\end{figure}

\begin{figure}[htbp]
    \centering
    \includegraphics[width=0.7\textwidth]{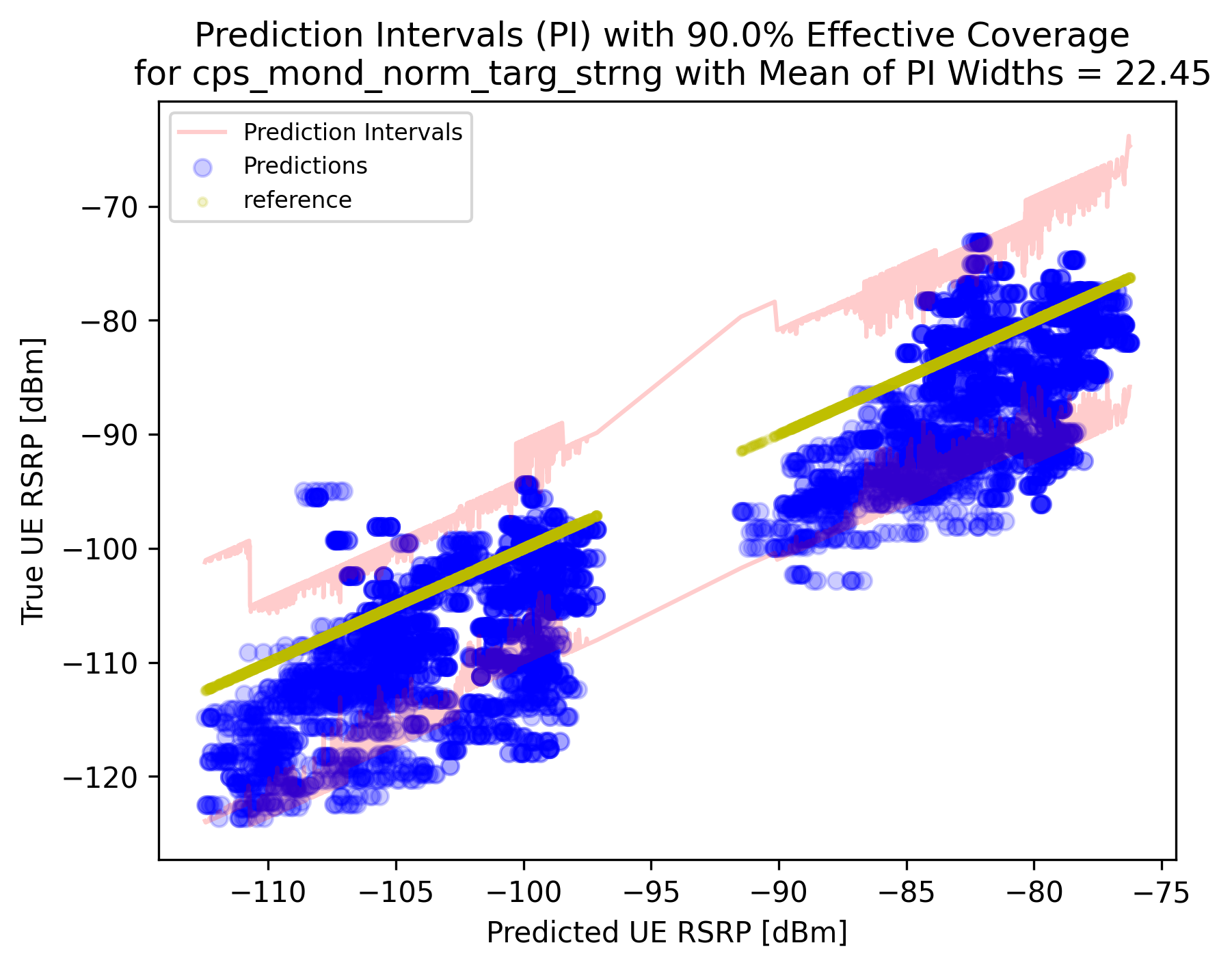}
    \caption{DTU RSRP PIs example: tabular-image model using internal tabular, image features}
\end{figure}

%Tabular-Image Model using Internal Tabular Feature
\begin{figure}[htbp]
    \centering
    \includegraphics[width=0.7\textwidth]{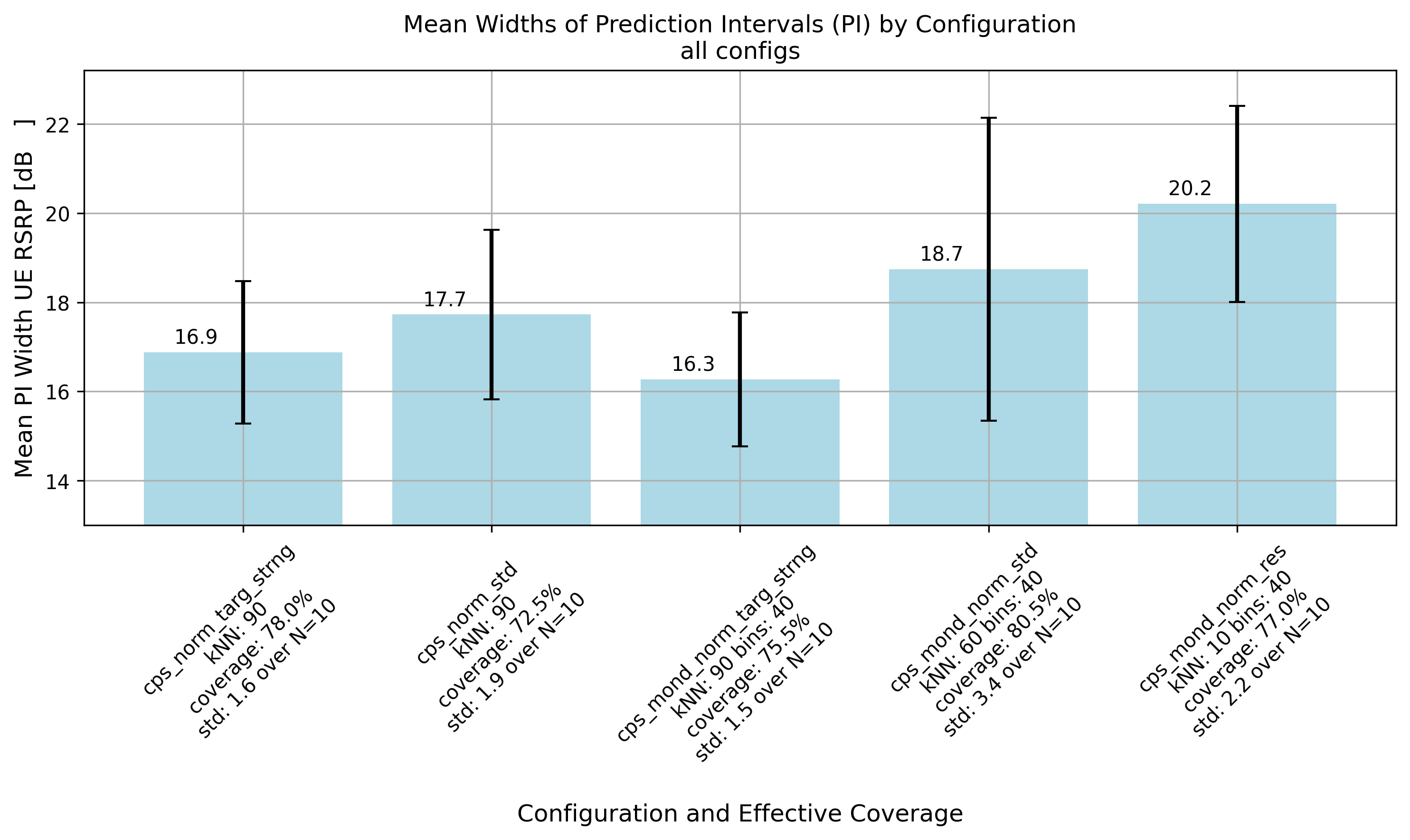}
    \caption{DTU RSRP all experiments: tabular-image model using internal tabular features}
\end{figure}

\begin{figure}[htbp]
    \centering
    \includegraphics[width=0.7\textwidth]{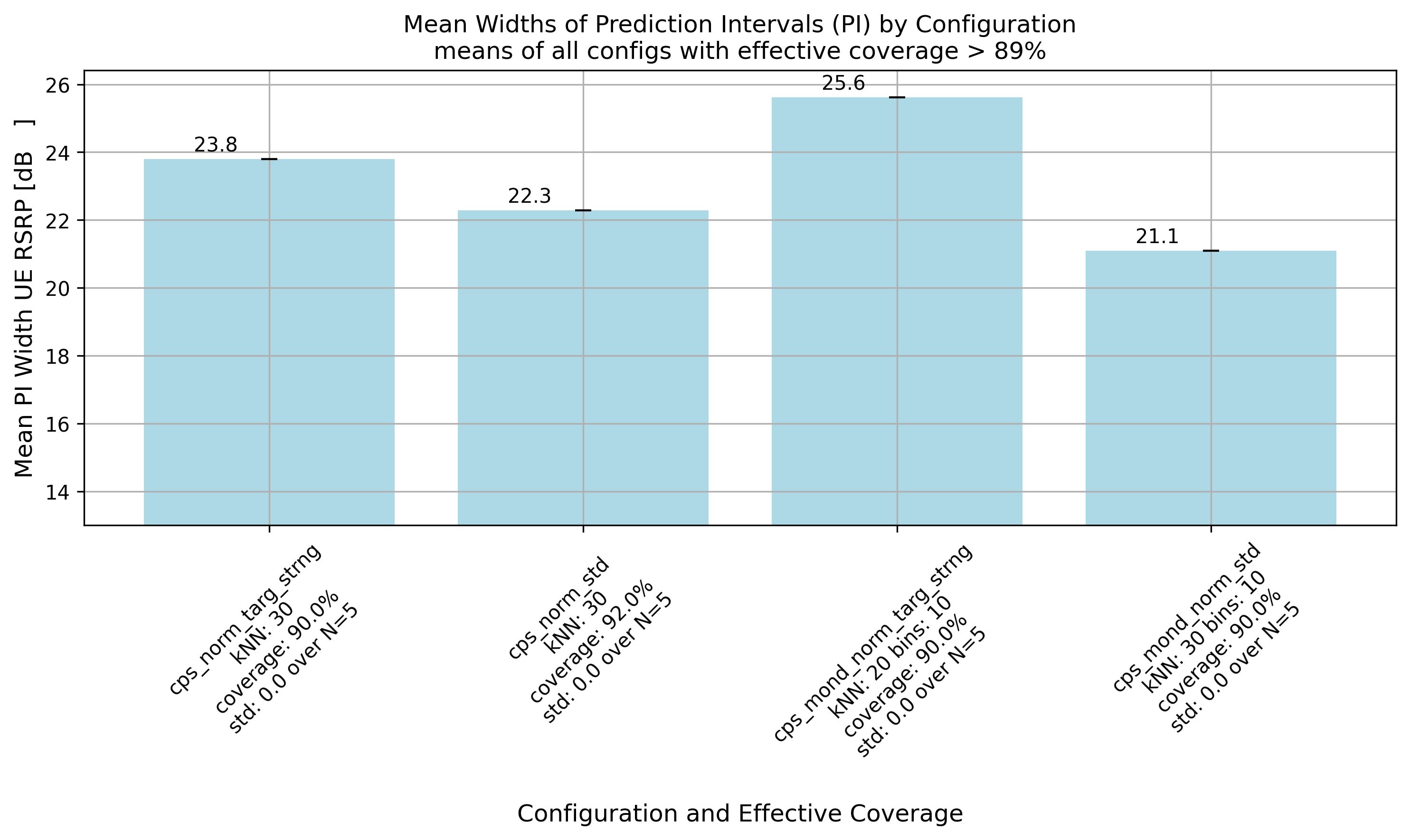}
    \caption{DTU RSRP experiments > 89\% coverage: tabular-image model using internal tabular feature }
\end{figure}

\begin{figure}[htbp]
    \centering
    \includegraphics[width=0.7\textwidth]{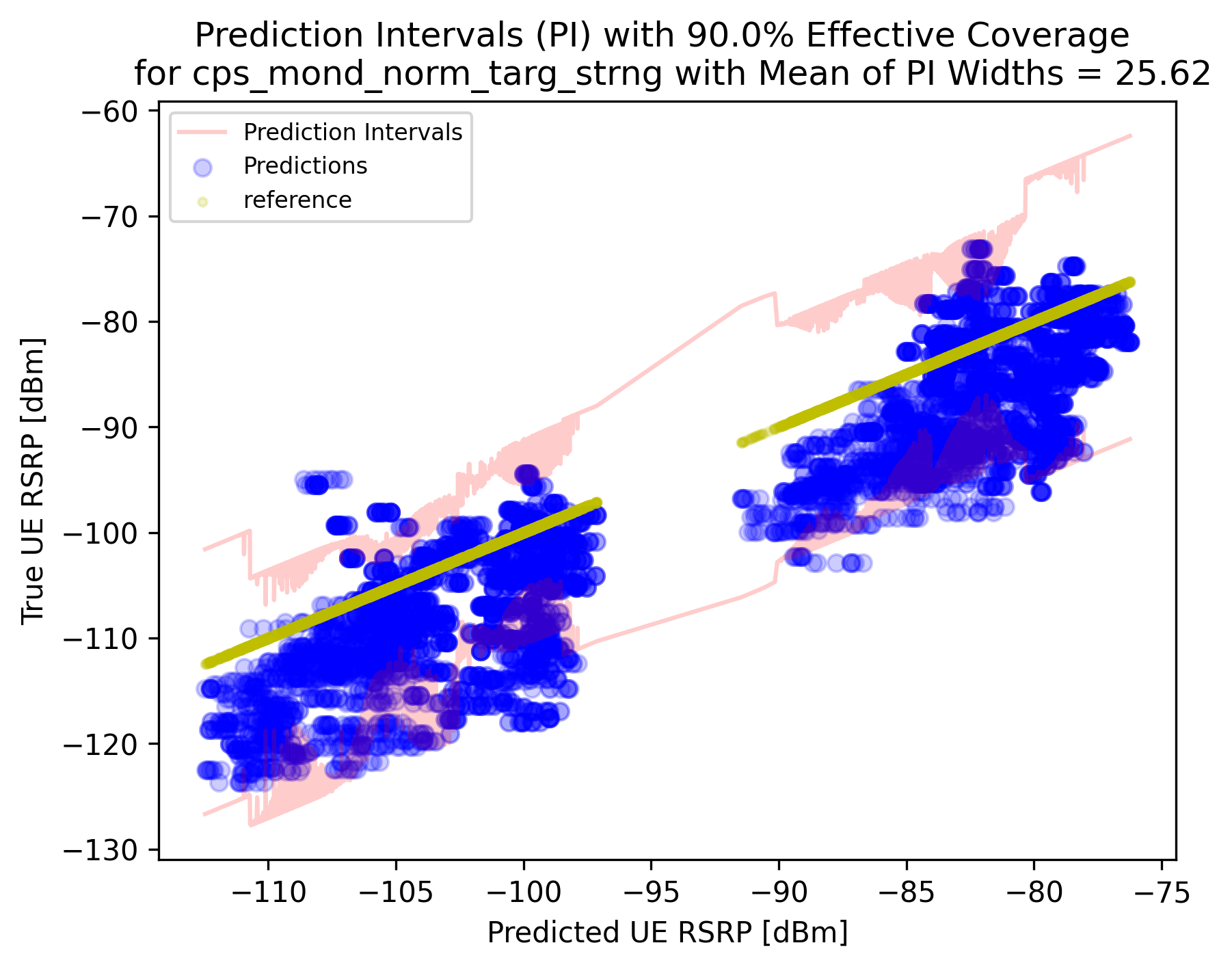}
    \caption{DTU RSRP PIs example: tabular-image model using internal tabular feature}
\end{figure}

%Tabular-Image Model using Internal Image Features
\begin{figure}[htbp]
    \centering
    \includegraphics[width=0.7\textwidth]{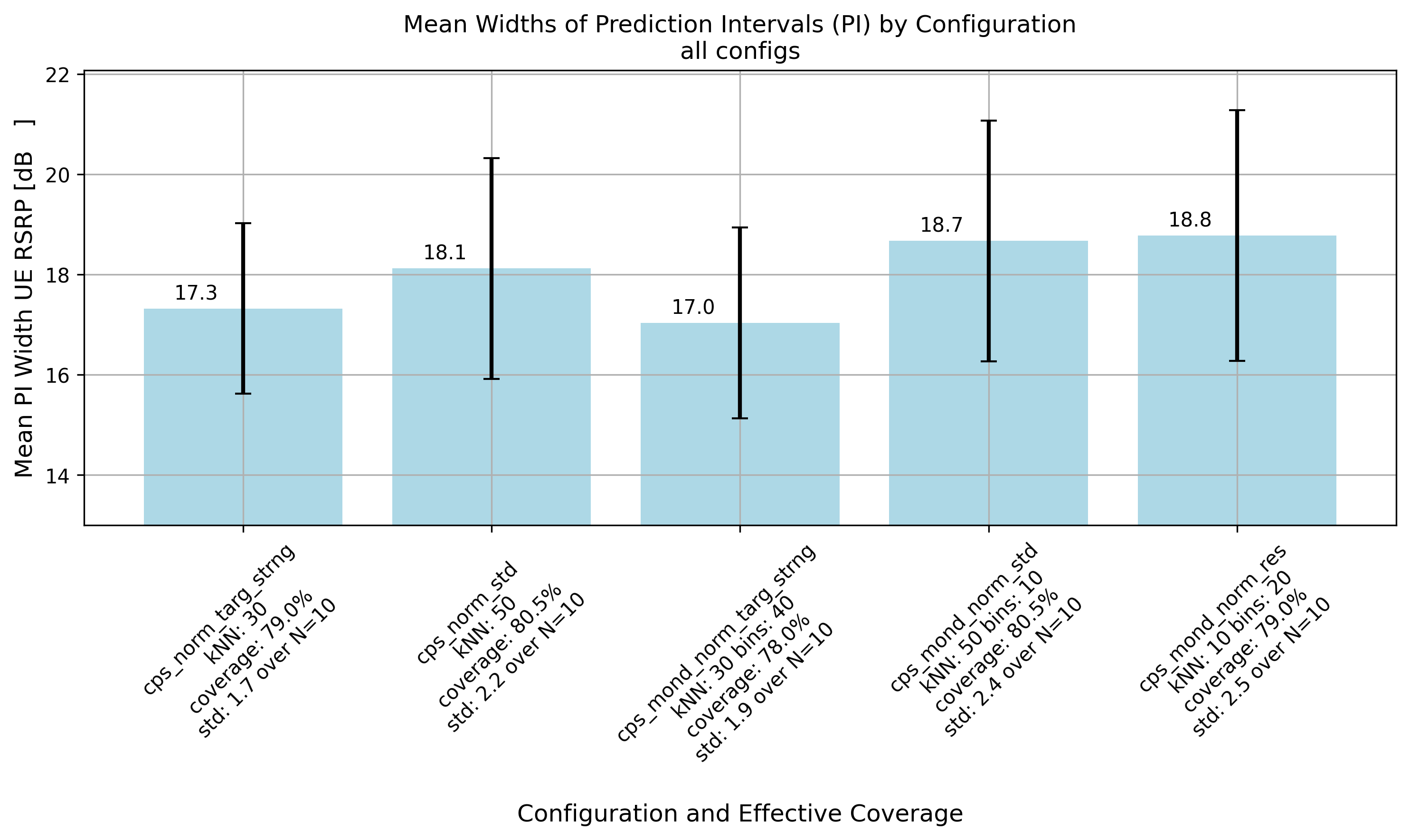}
    \caption{DTU RSRP all experiments: tabular-image model using internal image features}
\end{figure}

\begin{figure}[htbp]
    \centering
    \includegraphics[width=0.7\textwidth]{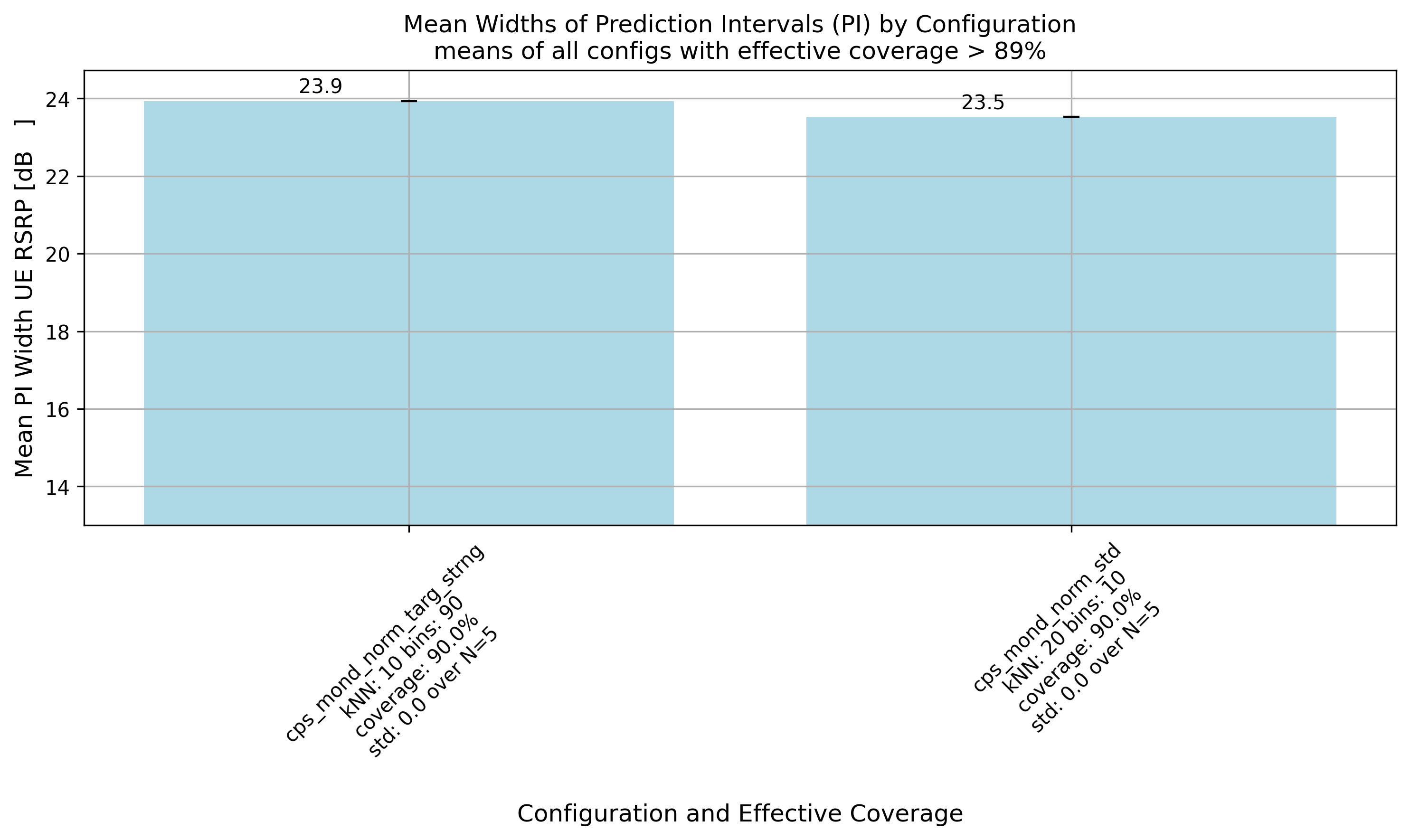}
    \caption{DTU RSRP experiments > 89\% coverage: tabular-image model using internal image features}
\end{figure}

\begin{figure}[htbp]
    \centering
    \includegraphics[width=0.7\textwidth]{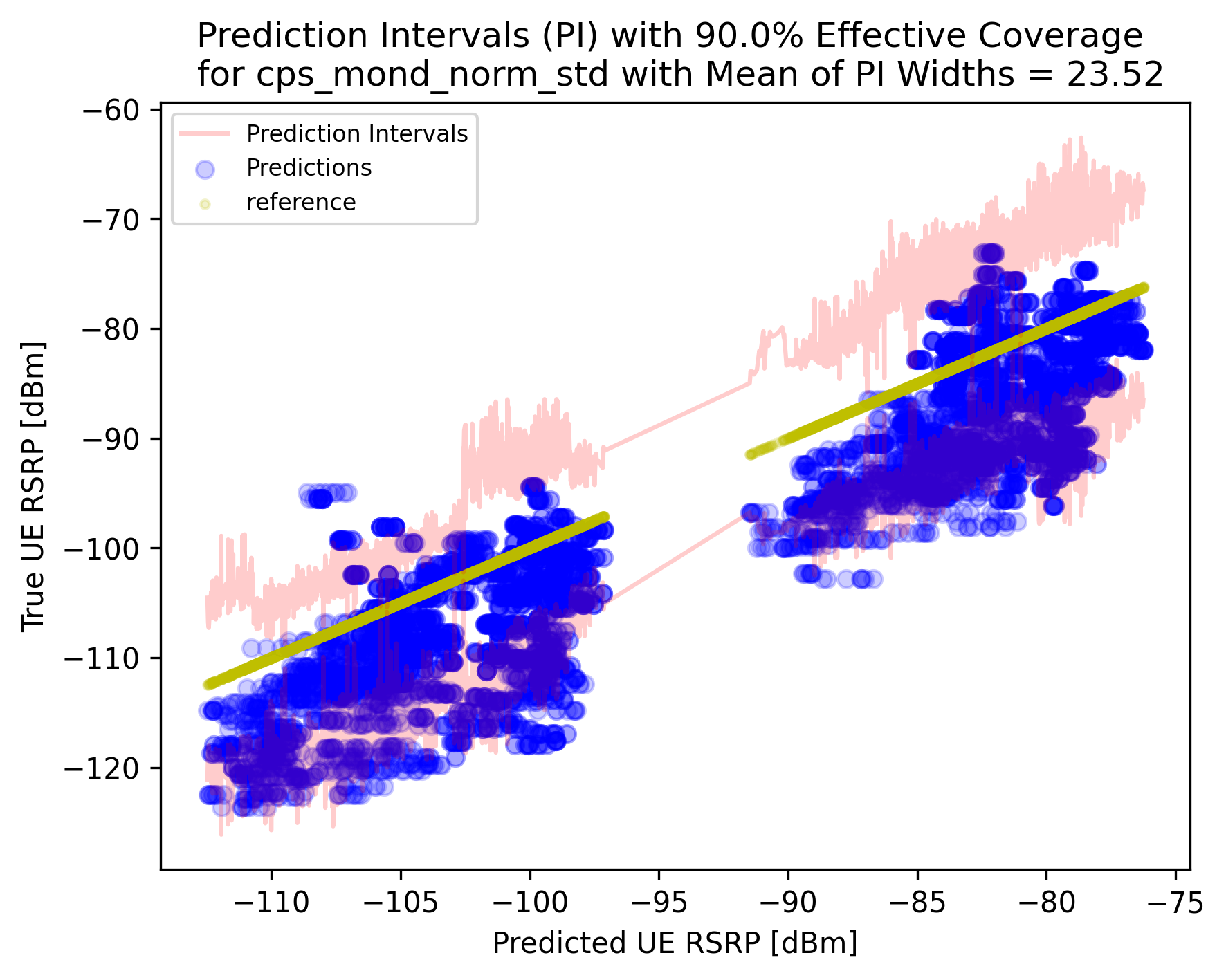}
    \caption{DTU RSRP PIs example: tabular-image model using internal image features}
\end{figure}

%Image-Only Model using Internal Image Features
\begin{figure}[htbp]
    \centering
    \includegraphics[width=0.7\textwidth]{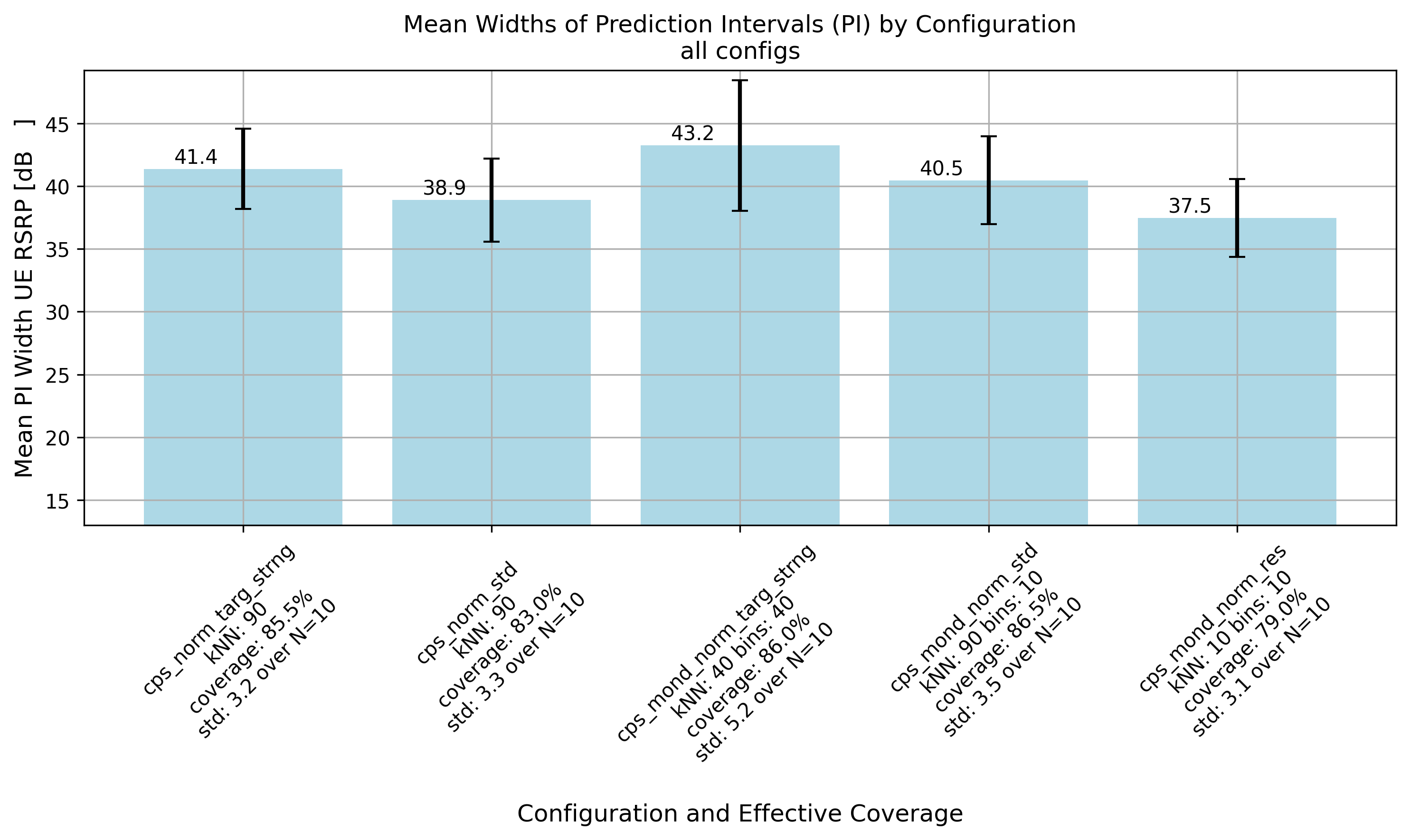}
    \caption{DTU RSRP all experiments: image-only model using internal image features}
\end{figure}

\begin{figure}[htbp]
    \centering
    \includegraphics[width=0.7\textwidth]{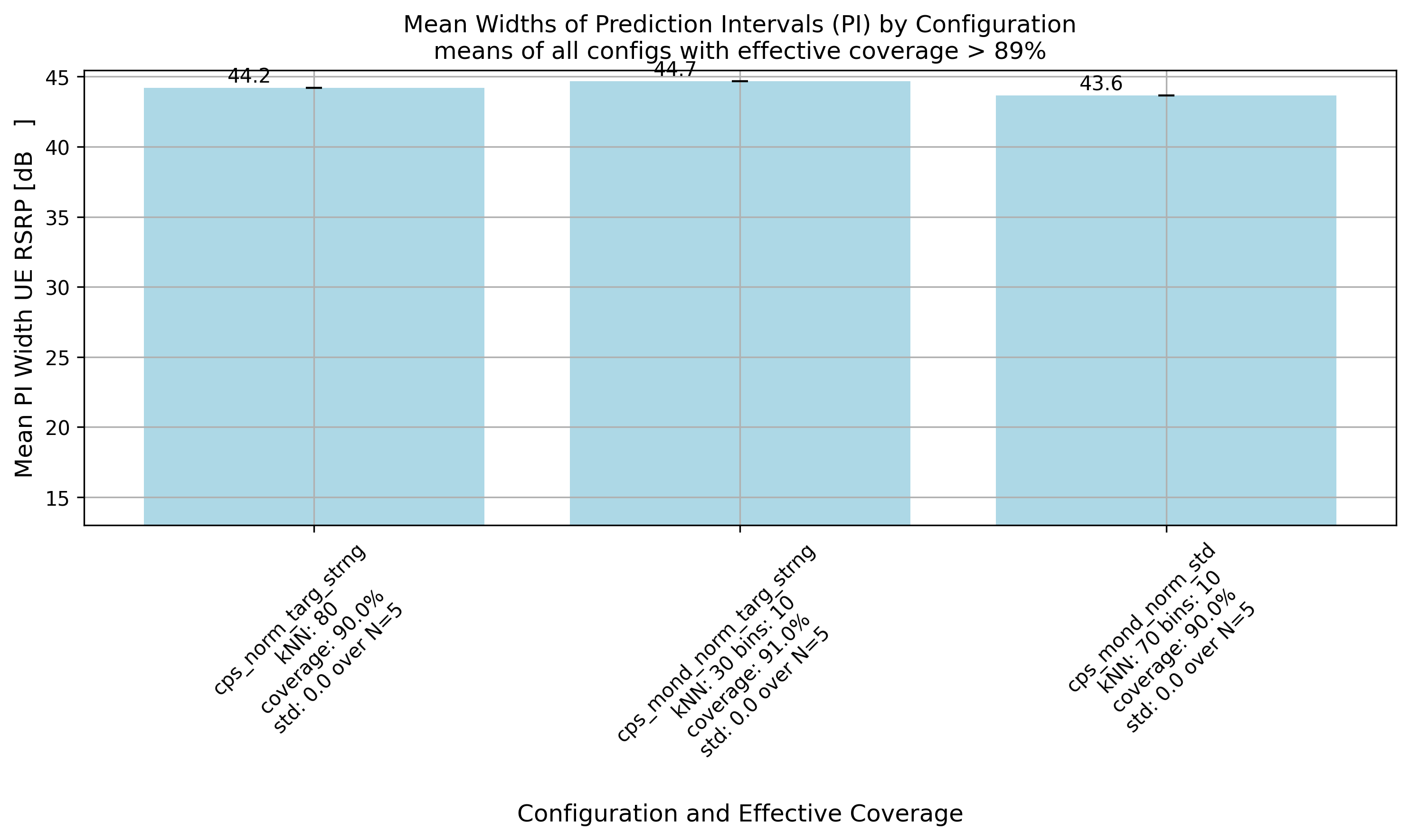}
    \caption{DTU RSRP experiments > 89\% coverage: image-only model using internal image features}
\end{figure}

\begin{figure}[htbp]
    \centering
    \includegraphics[width=0.7\textwidth]{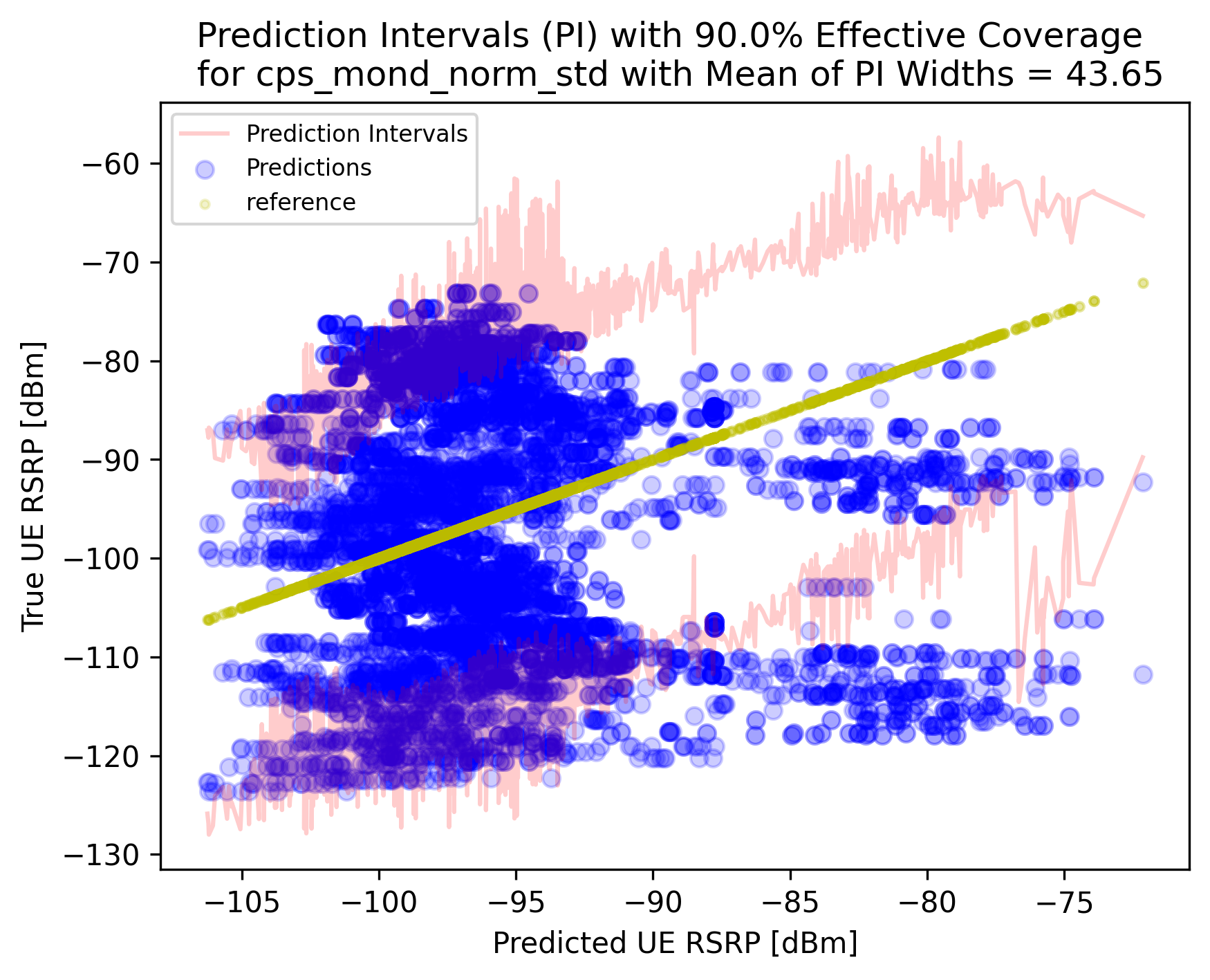}
    \caption{DTU RSRP PIs example: image-only model using internal image features}
\end{figure}
\clearpage
\newpage
\subsubsection{Multimodal Toolkit: price regression PIs results}

%External Numerical Features
\begin{figure}[htbp]
    \centering
    \includegraphics[width=0.7\textwidth]{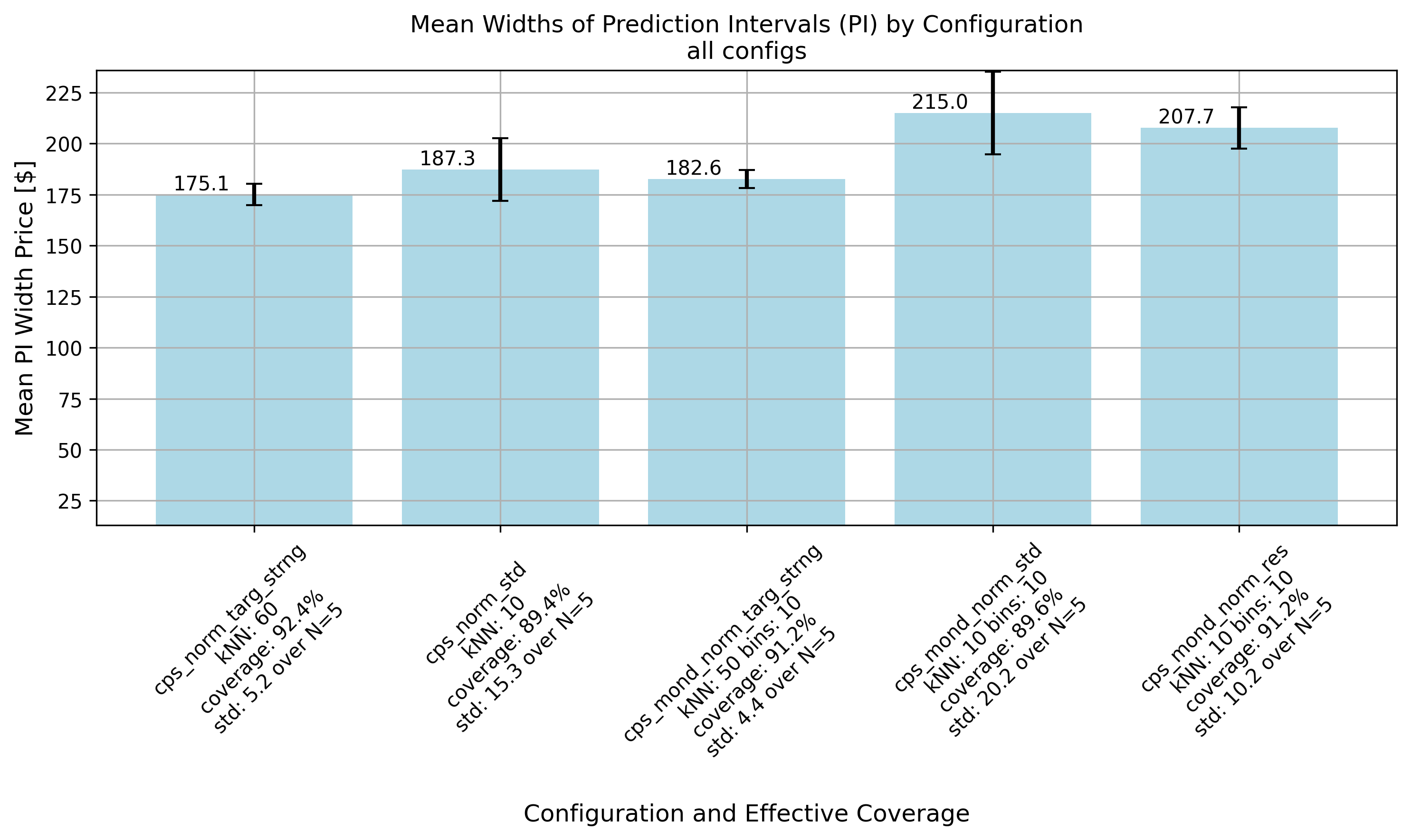}
    \caption{Multimodal price all experiments: external numerical features}
\end{figure}

\begin{figure}[htbp]
    \centering
    \includegraphics[width=0.7\textwidth]{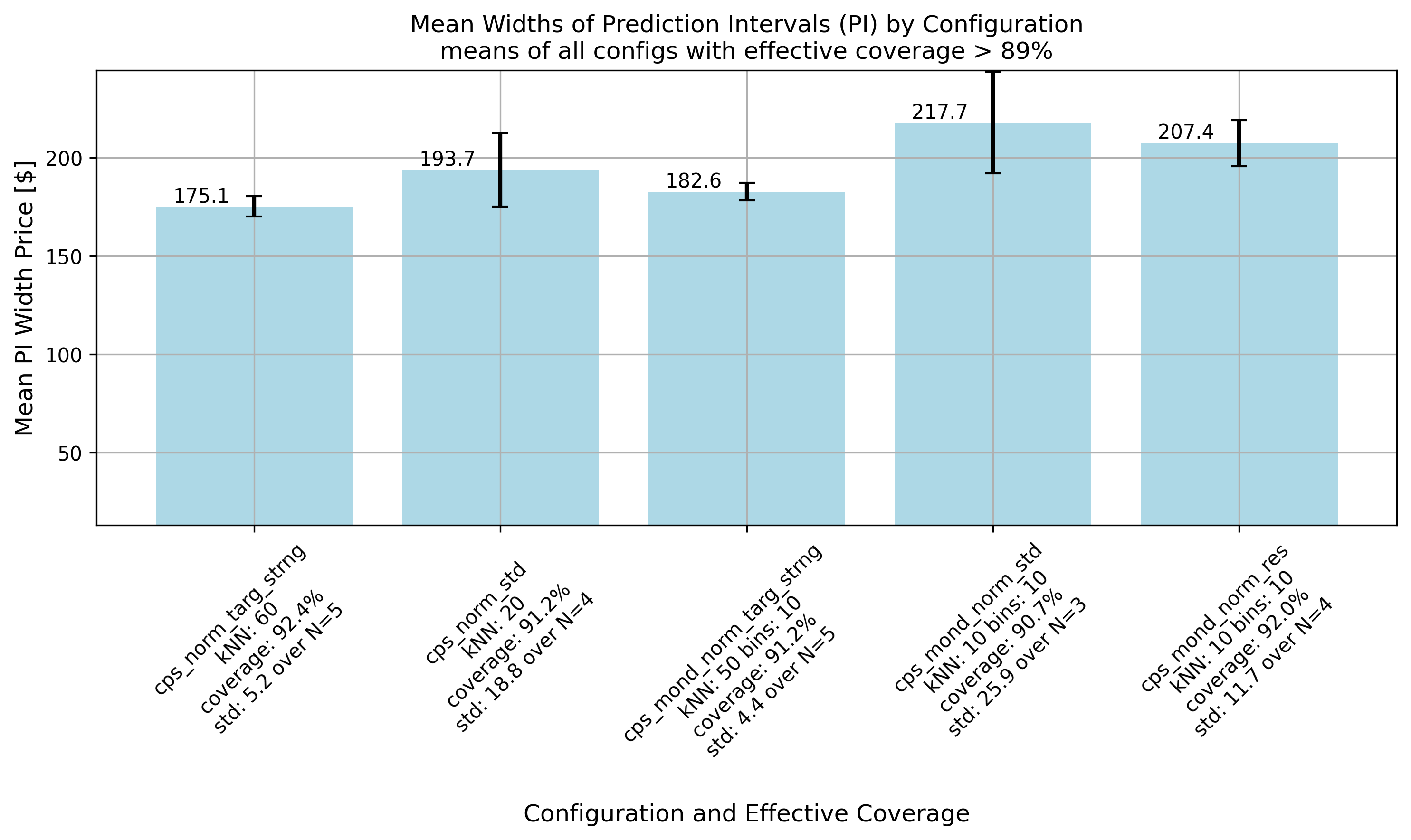}
    \caption{Multimodal price experiments > 89\% coverage: external numerical features}
\end{figure}

\begin{figure}[htbp]
    \centering
    \includegraphics[width=0.7\textwidth]{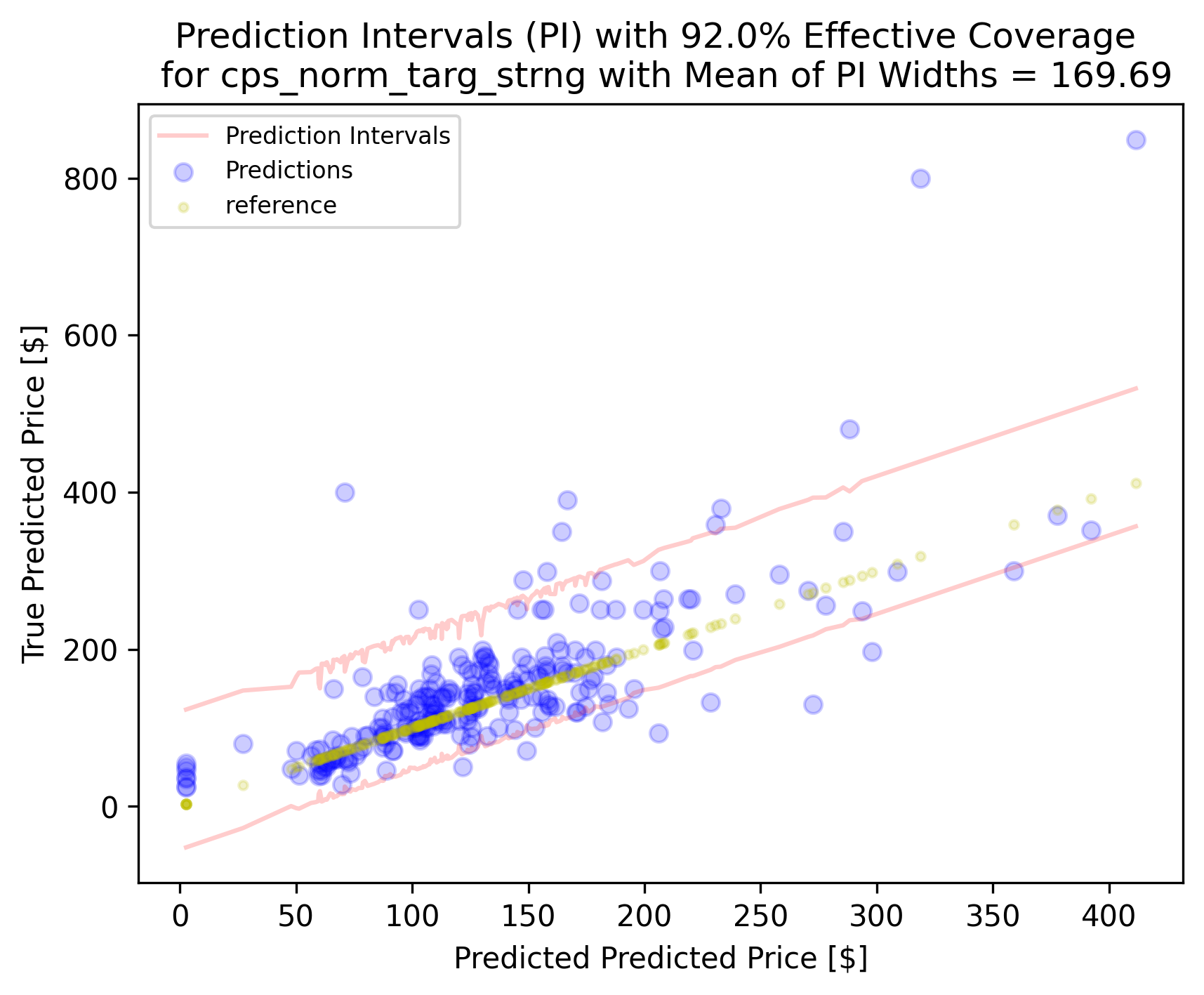}
    \caption{Multimodal price PIs example: external numerical features}
\end{figure}

%Internal BERT, Categorical, Numerical, and Feature
\begin{figure}[htbp]
    \centering
    \includegraphics[width=0.7\textwidth]{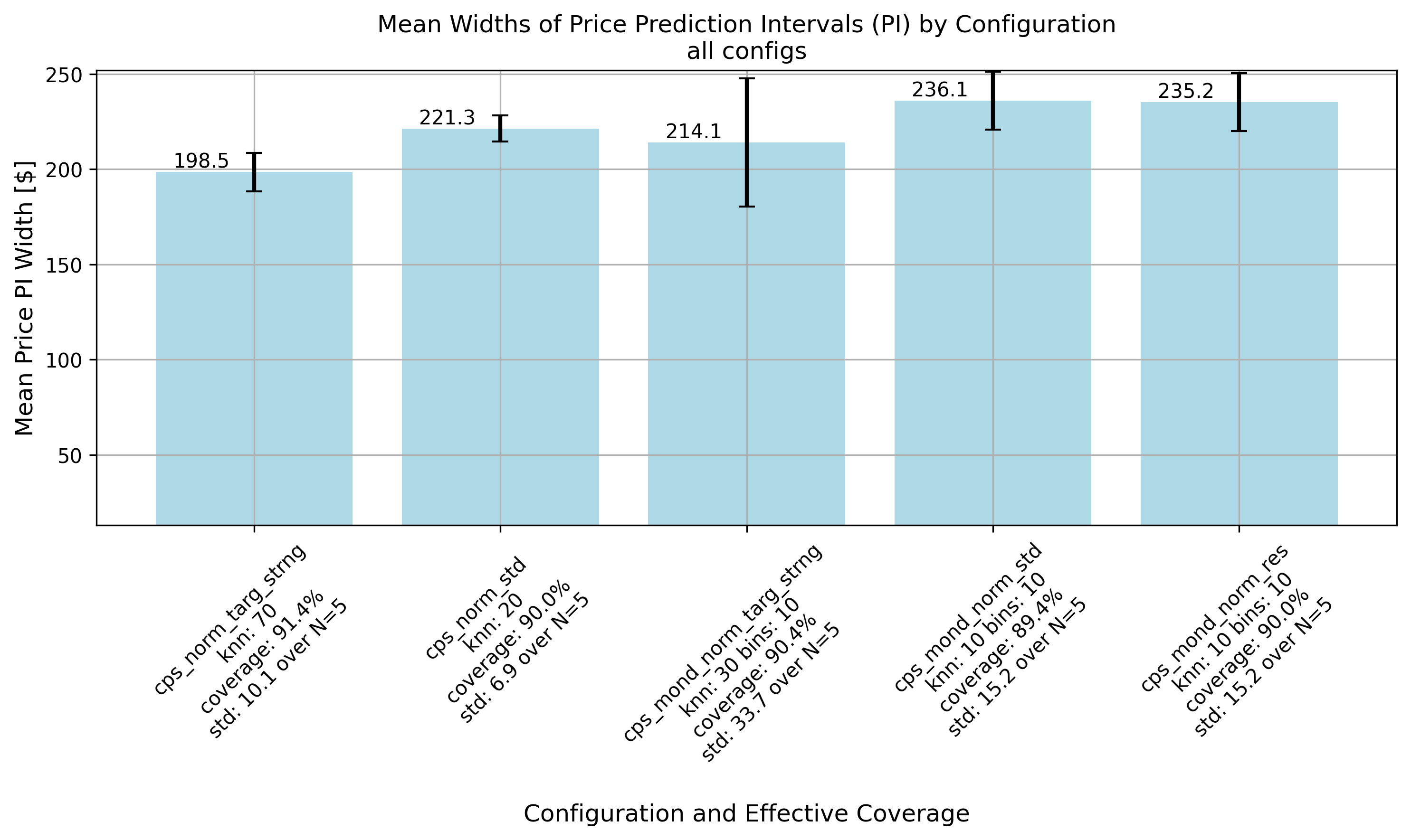}
    \caption{Multimodal price all experiments: internal BERT, categorical, and numerical features}
\end{figure}

\begin{figure}[htbp]
    \centering
    \includegraphics[width=0.7\textwidth]{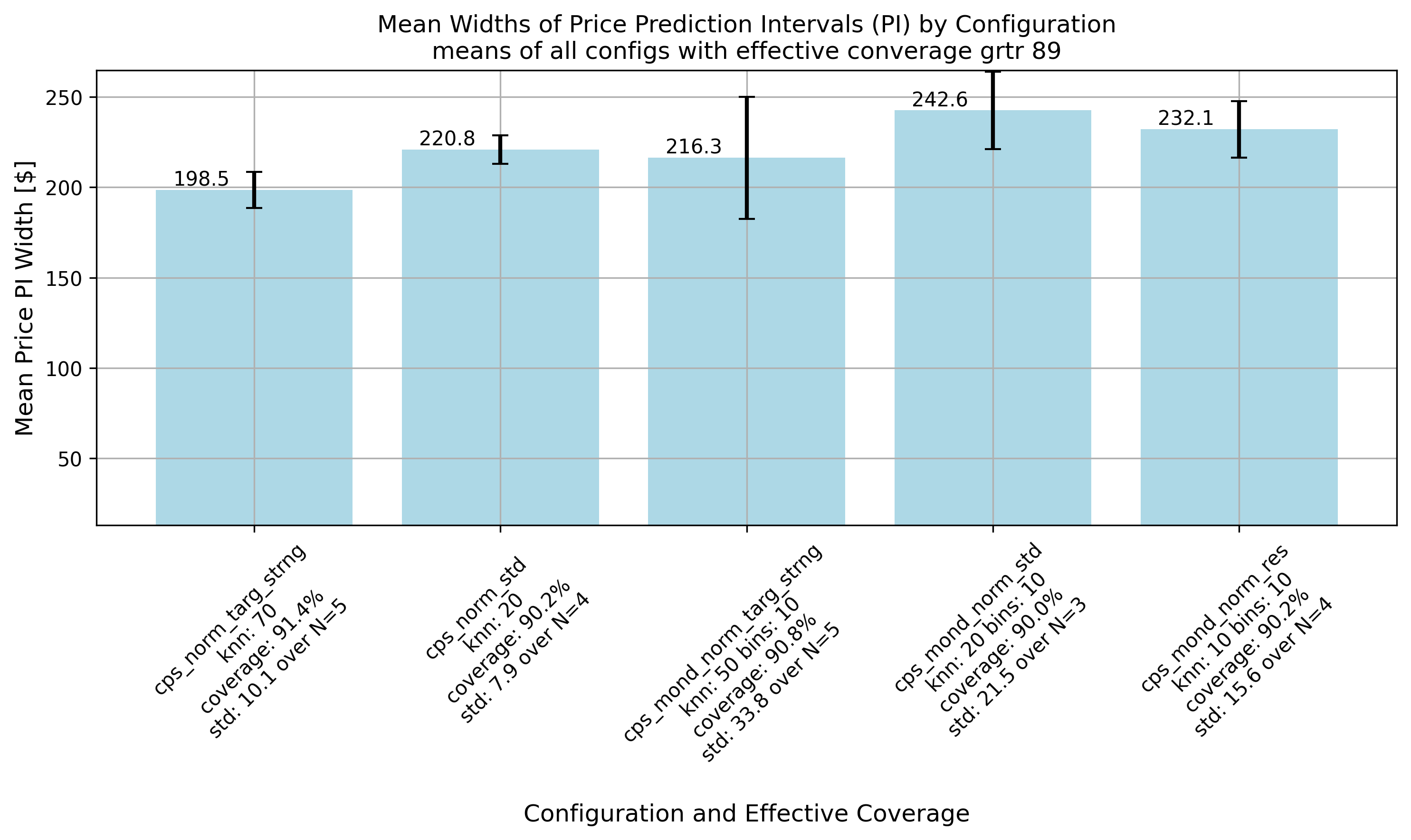}
    \caption{Multimodal price experiments > 89\% coverage: internal BERT, categorical, and numerical features}
\end{figure}

\begin{figure}[htbp]
    \centering
    \includegraphics[width=0.7\textwidth]{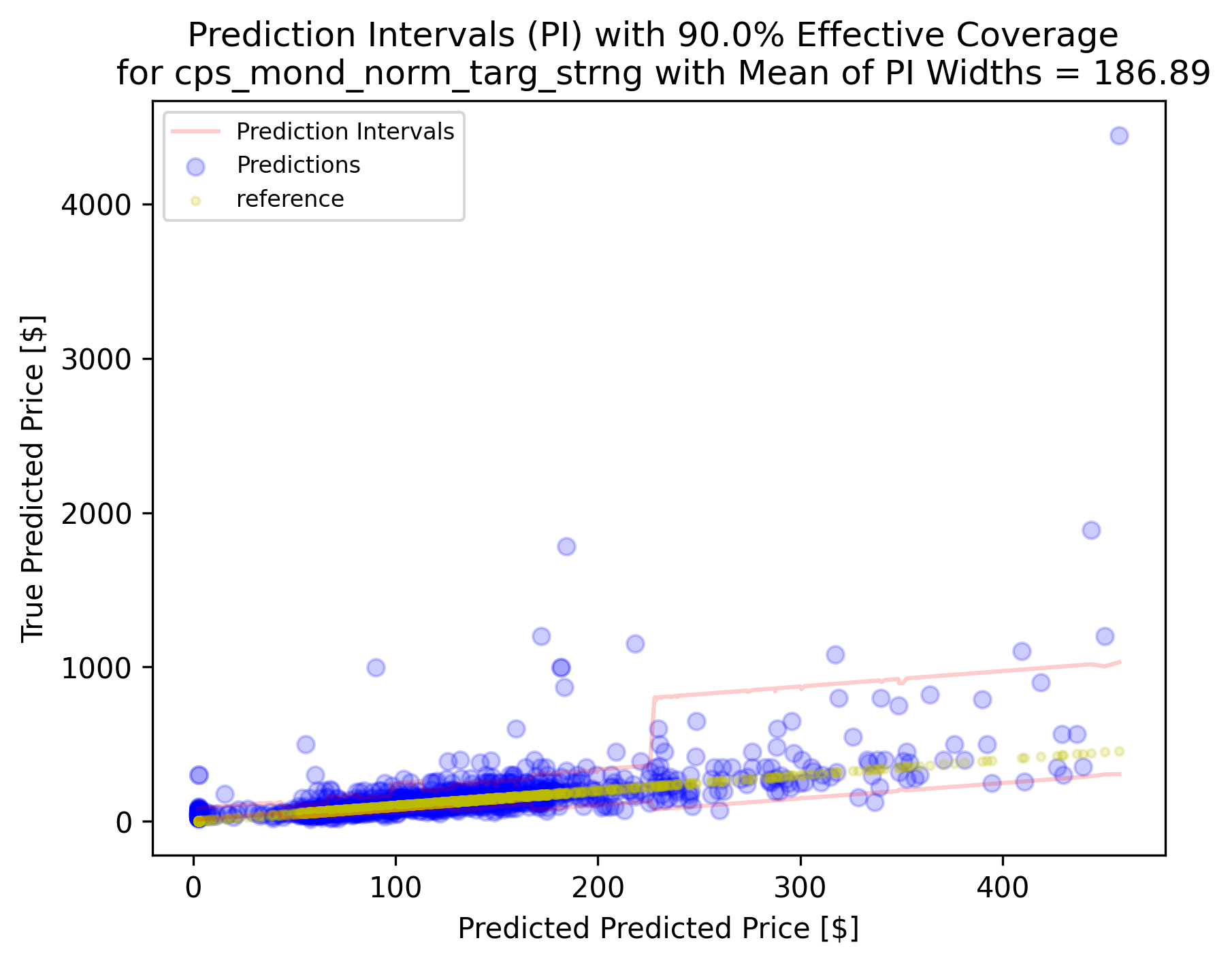}
    \caption{Multimodal price PIs example: internal BERT, categorical, and numerical features}
\end{figure}

%BERT-Only Features

\begin{figure}[htbp]
    \centering
    \includegraphics[width=0.7\textwidth]{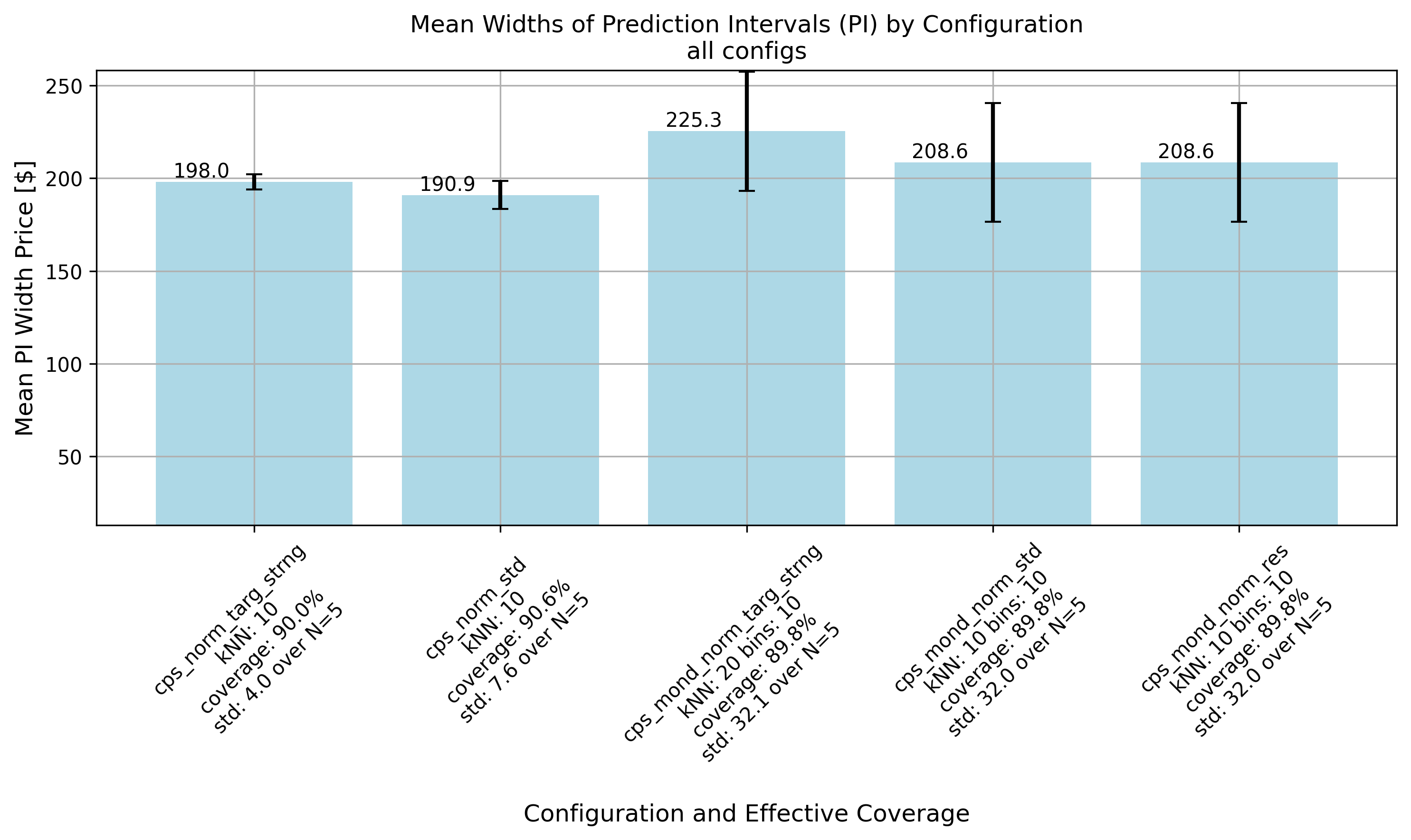}
    \caption{Multimodal price all experiments: BERT-only features}
\end{figure}

\begin{figure}[htbp]
    \centering
    \includegraphics[width=0.7\textwidth]{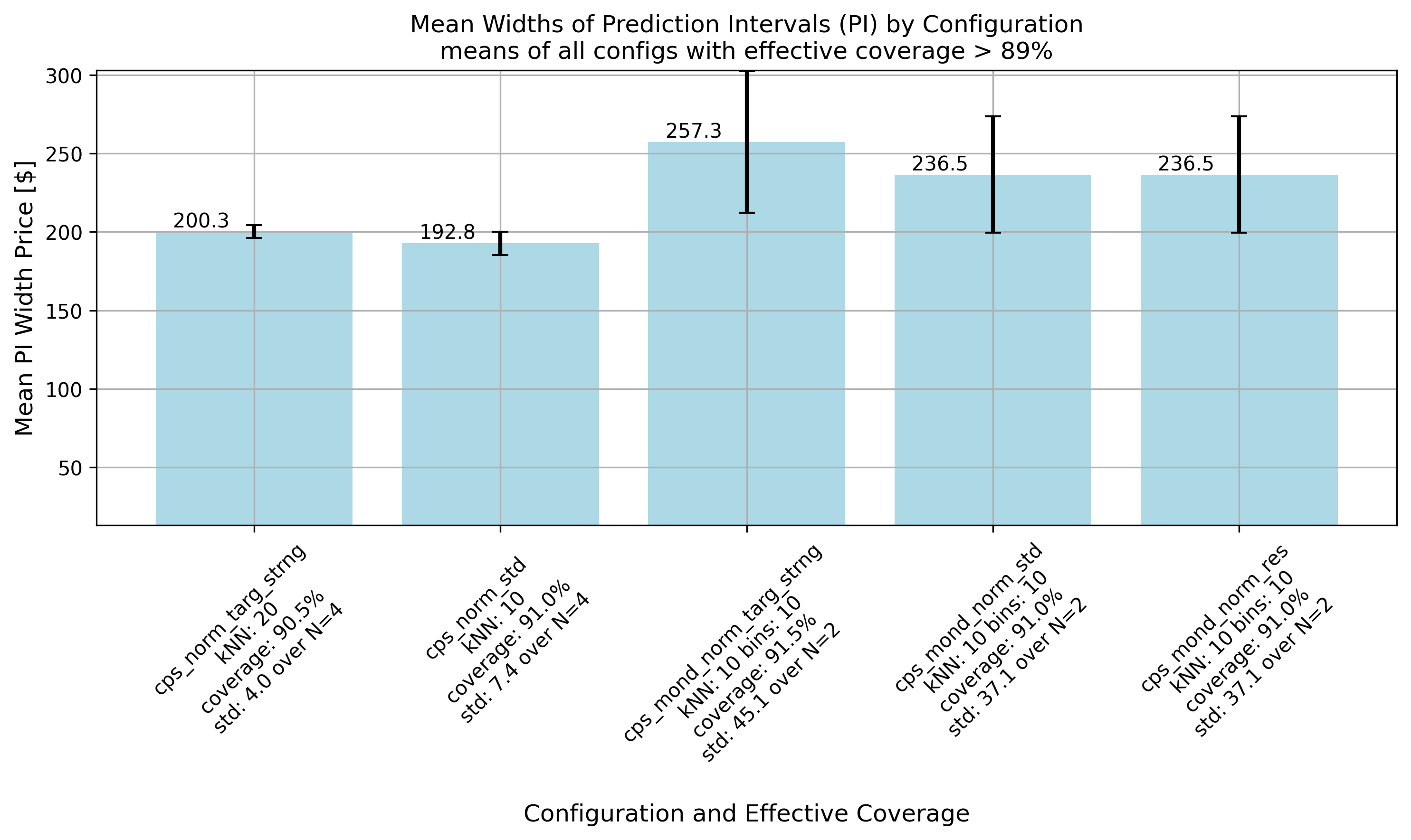}
    \caption{Multimodal price experiments > 89\% coverage: BERT-only features}
\end{figure}

\begin{figure}[htbp]
    \centering
    \includegraphics[width=0.7\textwidth]{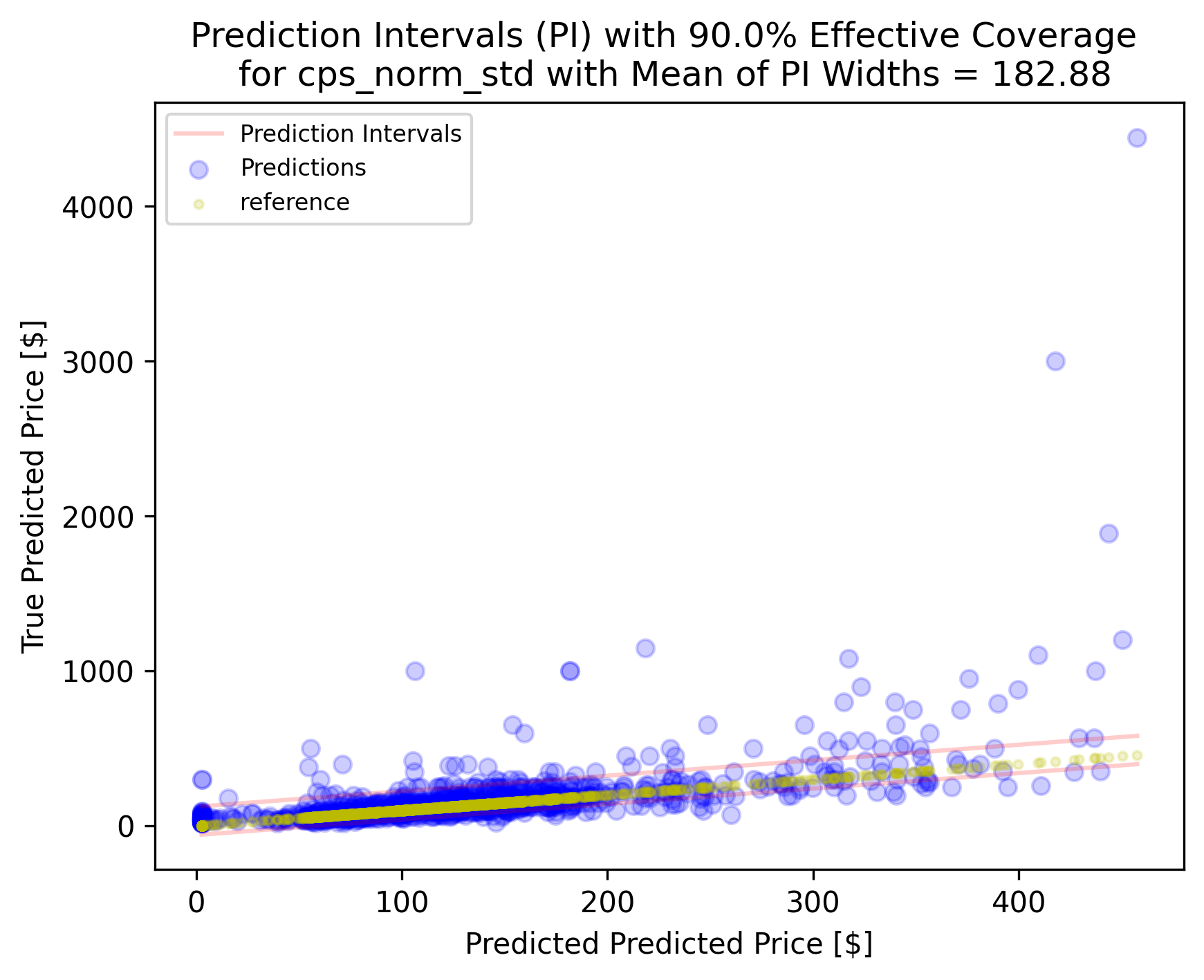}
    \caption{Multimodal price PIs example: BERT-Only}
\end{figure}

%Categorical-Only Features
\begin{figure}[htbp]
    \centering
    \includegraphics[width=0.7\textwidth]{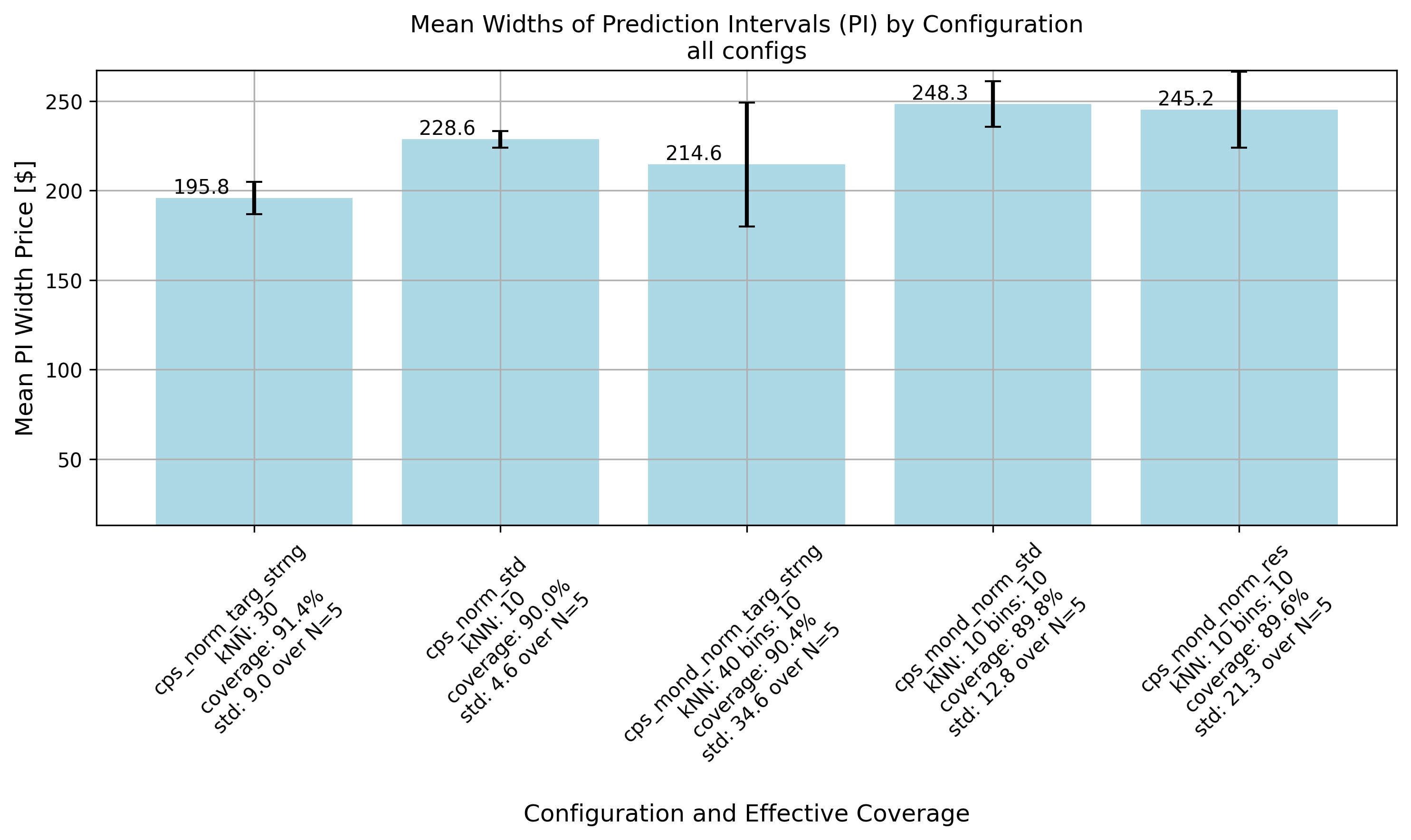}
    \caption{Multimodal price all experiments: internal categorical-only features}
\end{figure}

\begin{figure}[htbp]
    \centering
    \includegraphics[width=0.7\textwidth]{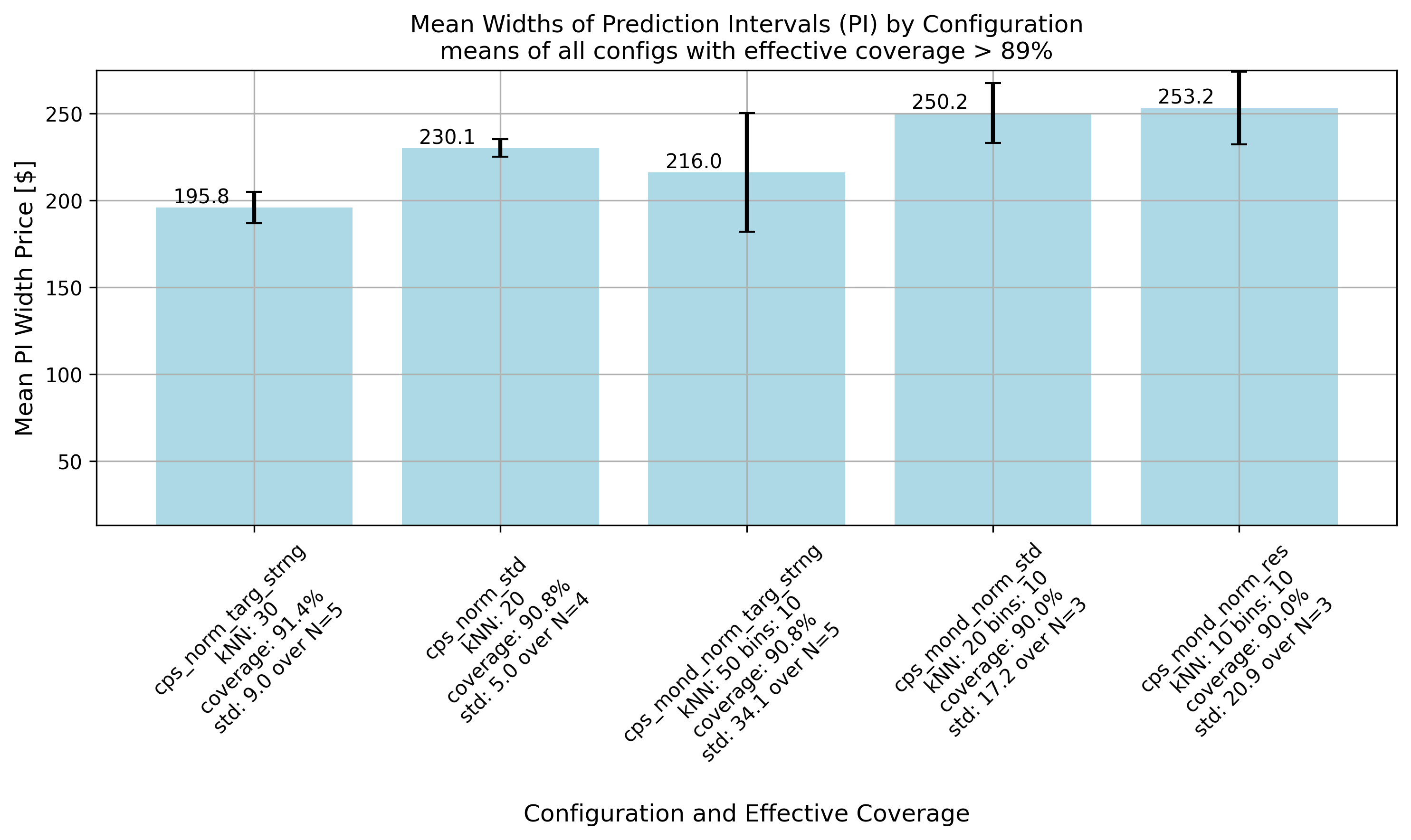}
    \caption{Multimodal price experiments > 89\% coverage: internal categorical-only features}
\end{figure}

\begin{figure}[htbp]
    \centering
    \includegraphics[width=0.7\textwidth]{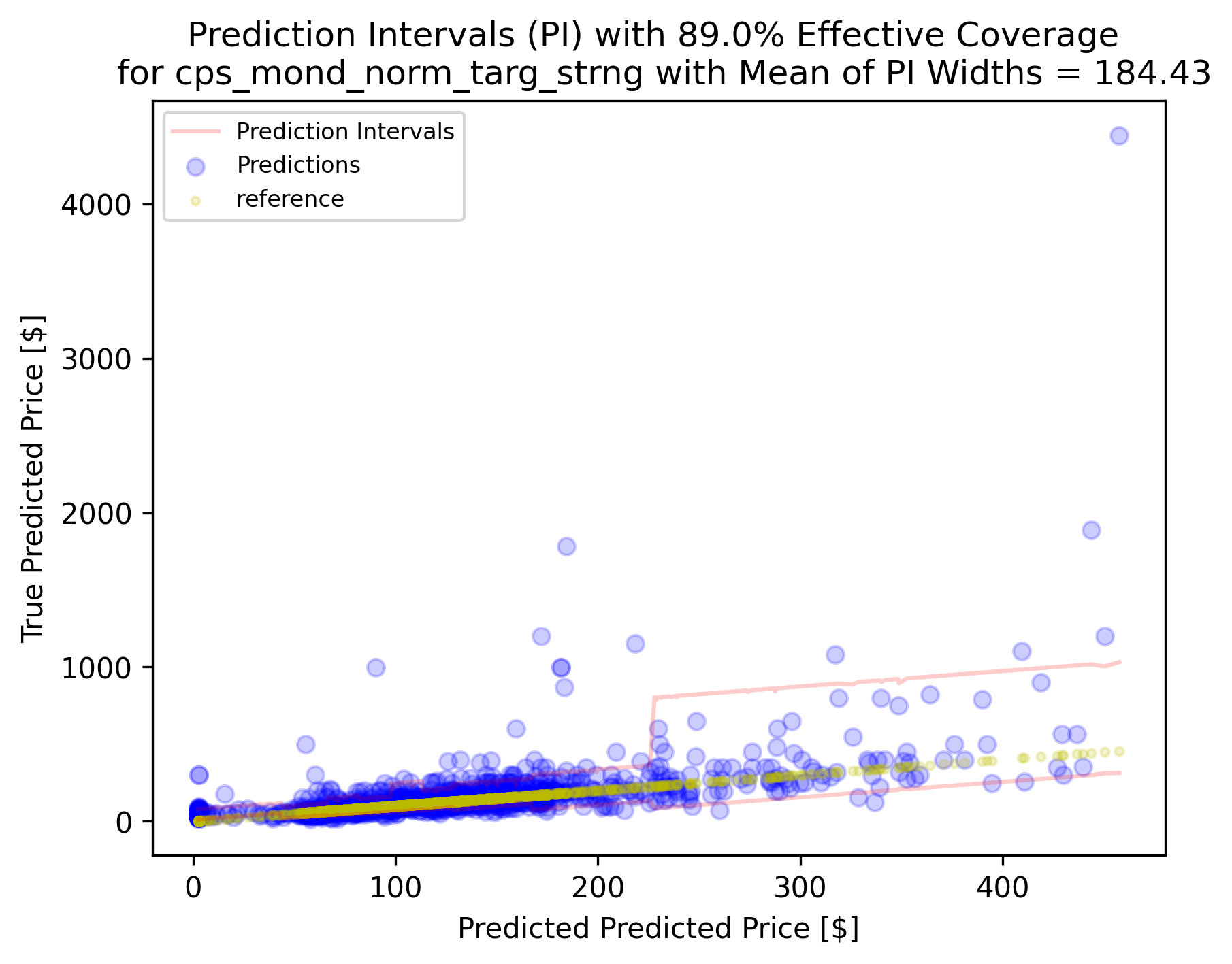}
    \caption{Multimodal price PIs example: internal categorical-only features}
\end{figure}

%Numerical-Only Features
\begin{figure}[htbp]
    \centering
    \includegraphics[width=0.7\textwidth]{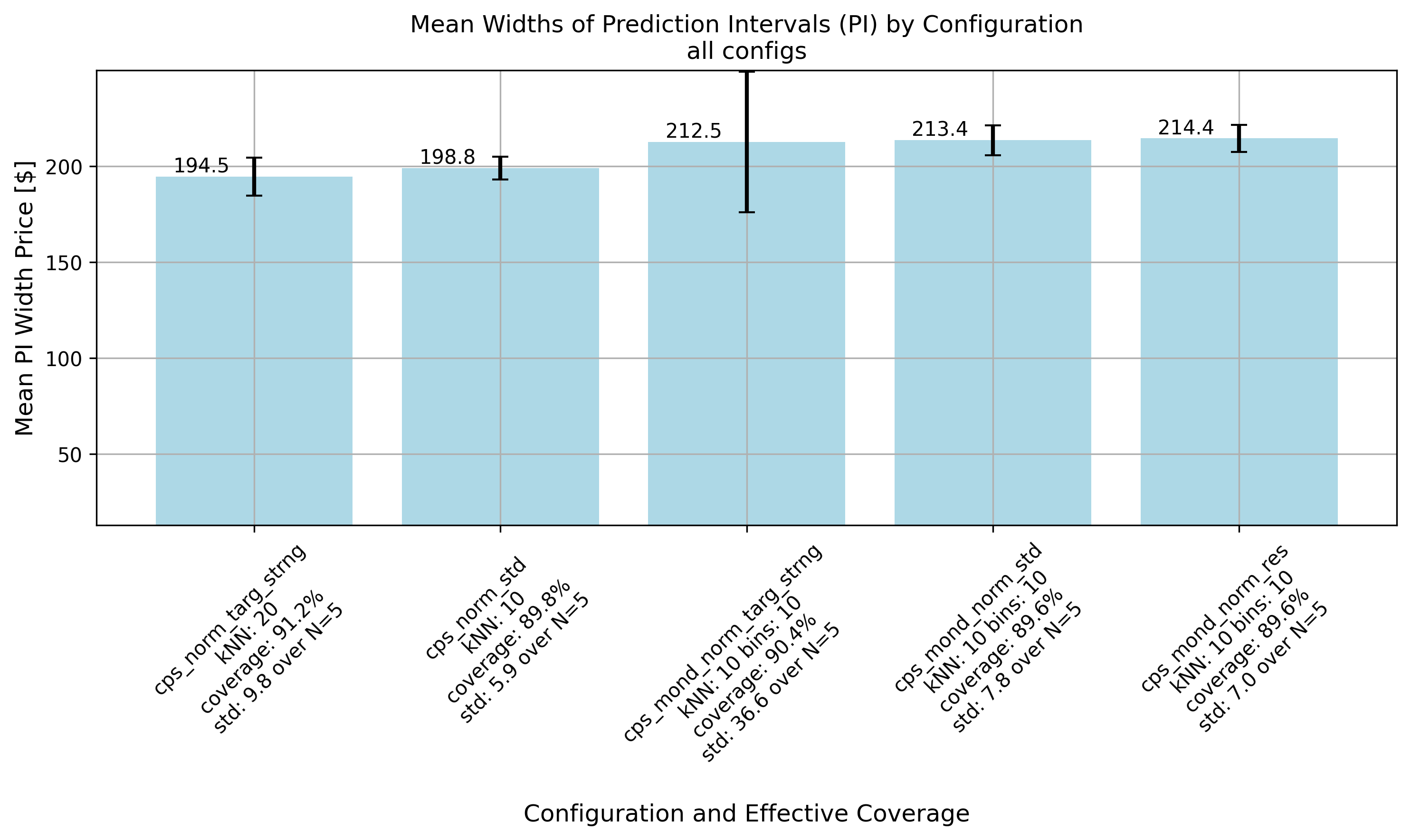}
    \caption{Multimodal price all experiments: internal numerical-only features}
\end{figure}

\begin{figure}[htbp]
    \centering
    \includegraphics[width=0.7\textwidth]{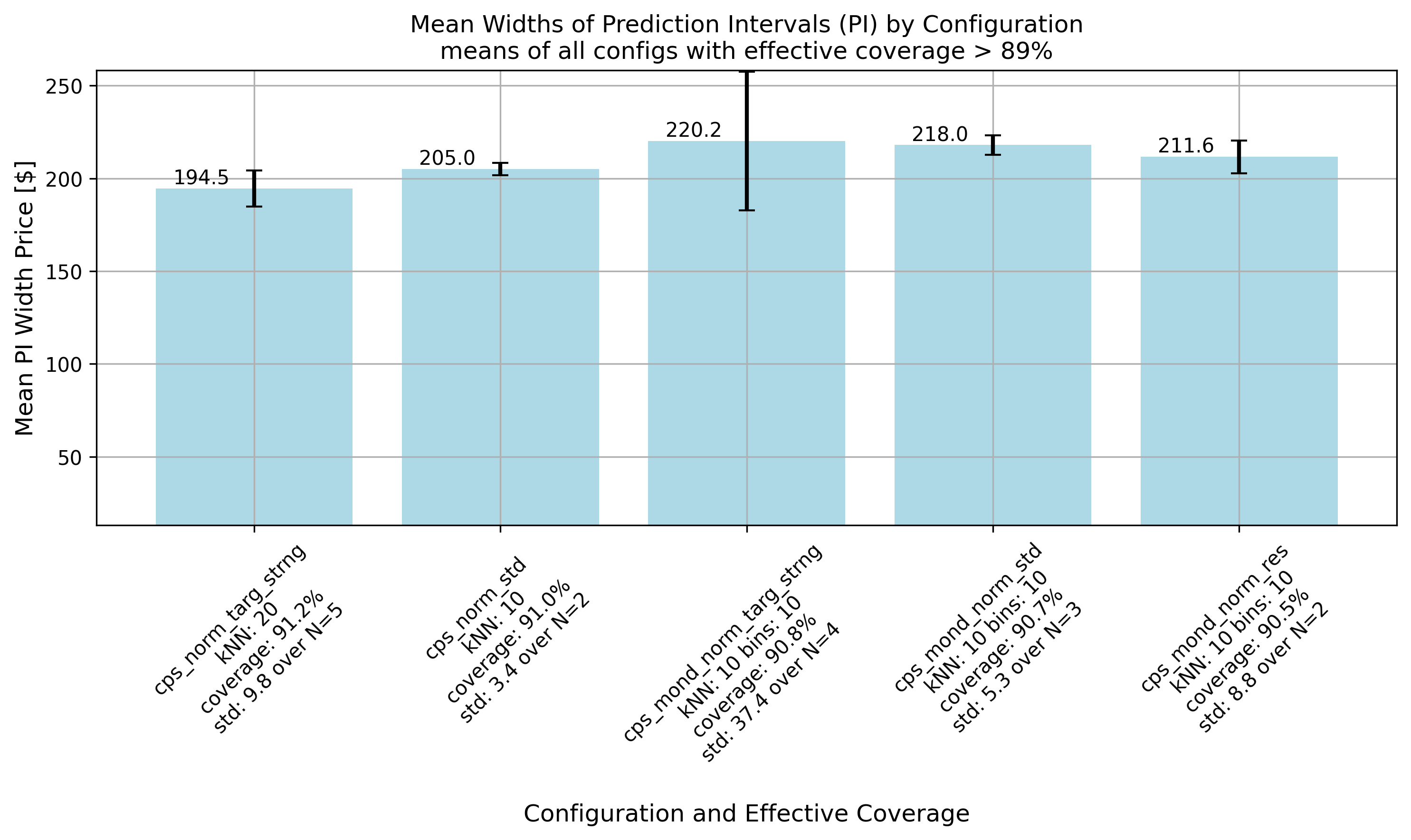}
    \caption{Multimodal price experiments > 89\% coverage: internal numerical-only features}
\end{figure}

\begin{figure}[htbp]
    \centering
    \includegraphics[width=0.7\textwidth]{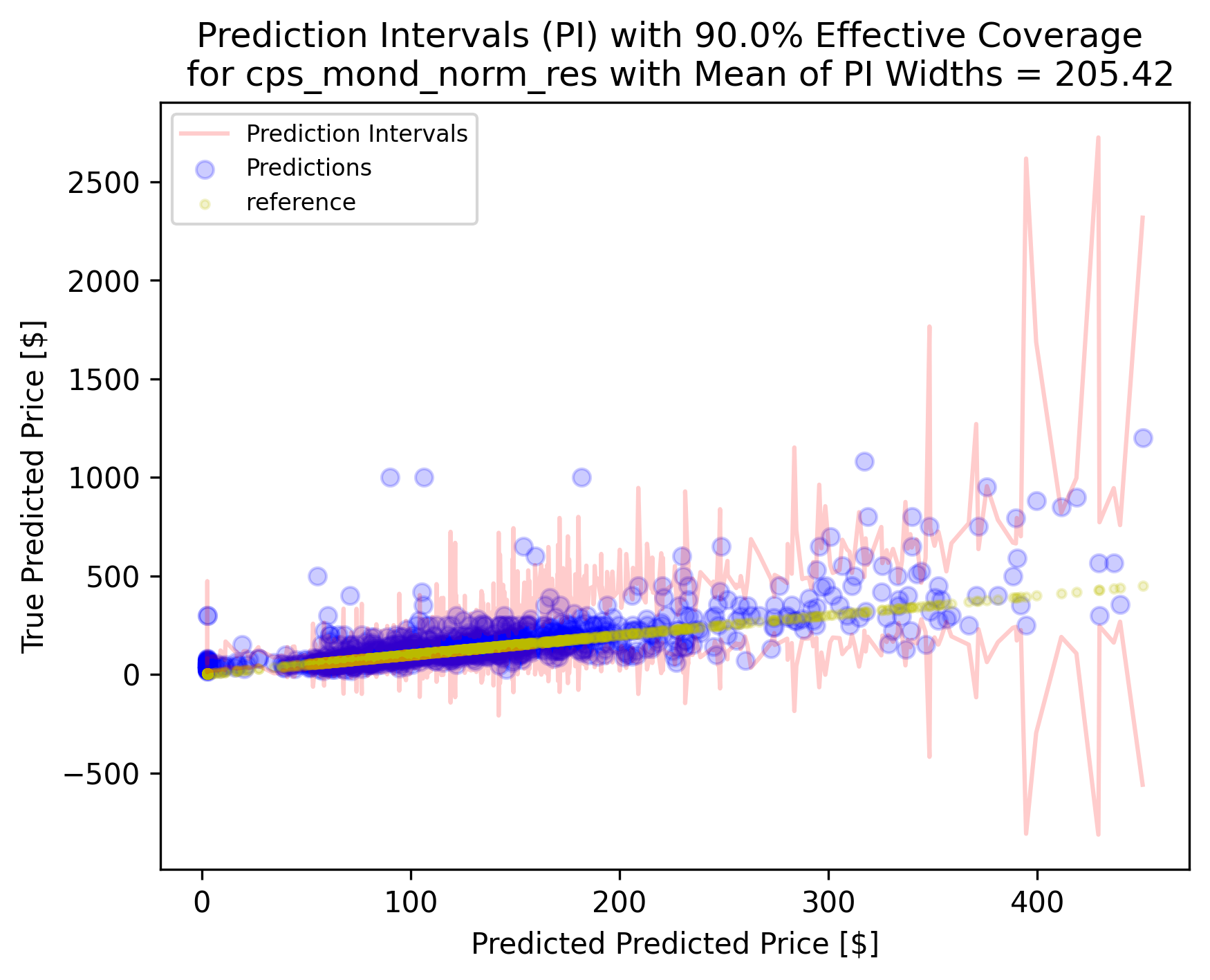}
    \caption{Multimodal price PIs example: internal numerical-only features}
\end{figure}

\end{document}